\documentclass[conference]{IEEEtran}
\usepackage{cite}

\ifCLASSINFOpdf
  \usepackage[pdftex]{graphicx}
  \DeclareGraphicsExtensions{.pdf,.jpeg,.png}
\else
\fi
\usepackage[cmex10]{amsmath}
\usepackage{amsfonts}

\usepackage{algorithmic}
\usepackage[tight,footnotesize]{subfigure}

\usepackage{float}
\usepackage{url}

\begin{document}
%
\title{Some Further Evidence about Magnification and Shape in Neural Gas}



%
\author{\IEEEauthorblockN{Giacomo Parigi\IEEEauthorrefmark{1},
Andrea Pedrini\IEEEauthorrefmark{2},
Marco Piastra\IEEEauthorrefmark{1}}
\IEEEauthorblockA{\IEEEauthorrefmark{1}Computer Vision and Multimedia Lab\\
Universit\`a degli Studi di Pavia,
Via Ferrata 1, 27100 Pavia, Italy}
\IEEEauthorblockA{\IEEEauthorrefmark{2}Dipartimento di Matematica ``Federigo Enriques''\\
Universit\`a degli Studi di Milano,
Via Cesare Saldini, 50, 20133 Milano, Italy}
}


\maketitle


\begin{abstract}
Neural gas (NG) is a robust vector quantization algorithm with a well-known mathematical model. According to this, the neural gas samples the underlying data distribution following a power law with a magnification exponent that depends on data dimensionality only. The effects of \emph{shape} in the input data distribution, however, are not entirely covered by the NG model above, due to the technical difficulties involved. The experimental work described here shows that \emph{shape} is indeed relevant in determining the overall NG behavior; in particular, some experiments reveal richer and complex behaviors induced by shape that cannot be explained by the power law alone.  Although a more comprehensive analytical model remains to be defined, the evidence collected in these experiments suggests that the NG algorithm has an interesting potential for detecting complex shapes in noisy datasets. 
\end{abstract}


%
\IEEEpeerreviewmaketitle

\section{Introduction}

Neural Gas (NG) \cite{martinetz1991neural} is a robust algorithm for vector quantization with a well-known analytical model that has been first described in \cite{Martinetz-etal93}\footnote{According to Google Scholar, \cite{Martinetz-etal93} has $1291$ citations.} and then expanded in other subsequent works \cite{villmann2009some, witoelar2008learning}. According to this model, the Neural Gas algorithm performs a \emph{stochastic gradient descent} over an energy function and converges to a final configuration whereby the density of NG units relates to the density of the underlying input data distribution via a power law that is usually called \emph{magnification} (see e.g. \cite{villmann2006magnification}). Further studies have been conducted about the \emph{topographic mapping} properties of the NG \cite{Martinetz-Schulten94}, namely the capability to represent an implicit neighborhood relation -- which can be made explicit via Delaunay triangulation -- that describes faithfully the topology of input data distributions. Other studies focused on NG variants that enable controlling the power law explicitly \cite{jain2004forbidden, claussen2005magnification, villmann2006magnification, hammer2007magnification}, in particular to obtain an exponent of value one that makes the two densities be the same, except for a scaling factor.

The experimental study presented here is motivated by the objective of using NG for \emph{geometric inference} and \emph{shape detection}, as it happens with the NG-related algorithm SOAM \cite{piastra:soam}, that reconstructs a surface from a scattered point cloud of samples with topological and geometrical guarantees of faithfulness of the result obtained. For instance, this algorithm has been used for excluded volume computations in \cite{piastra2013octupolar}. More precisely, the focus of the study presented here is assessing the effects of \emph{shape} in input data distributions over the NG behavior.  For doing this, we consider input data distributions of the kind
$$\mathcal{U}(\text{\emph{shape}}) \ast N(\mathbf{0},\mathbf{\theta})$$
namely a uniform probability over a given \emph{shape} convolved with some (isotropic) noise with specific parameters $\mathbf{\theta}$.

Clearly, uniform shape sampling and isotropic noise are indeed limited representations of a real-world scenario for practical applications (e.g. for 2D or 3D image deconvolution). Nevertheless, even within these limits, the experimental evidence collected reveals the substantial relevance of shape in determining the overall NG behavior and shows an interesting potential in this direction; as it will be discussed, while the relevance of power law is substantially confirmed, the effects of shape in data distributions can be even stronger in some circumstances and determine NG configurations that cannot be explained by the power law alone.
 
\begin{figure}[!hb]
  \centering
  \subfigure[]{\includegraphics[width=.49\linewidth]{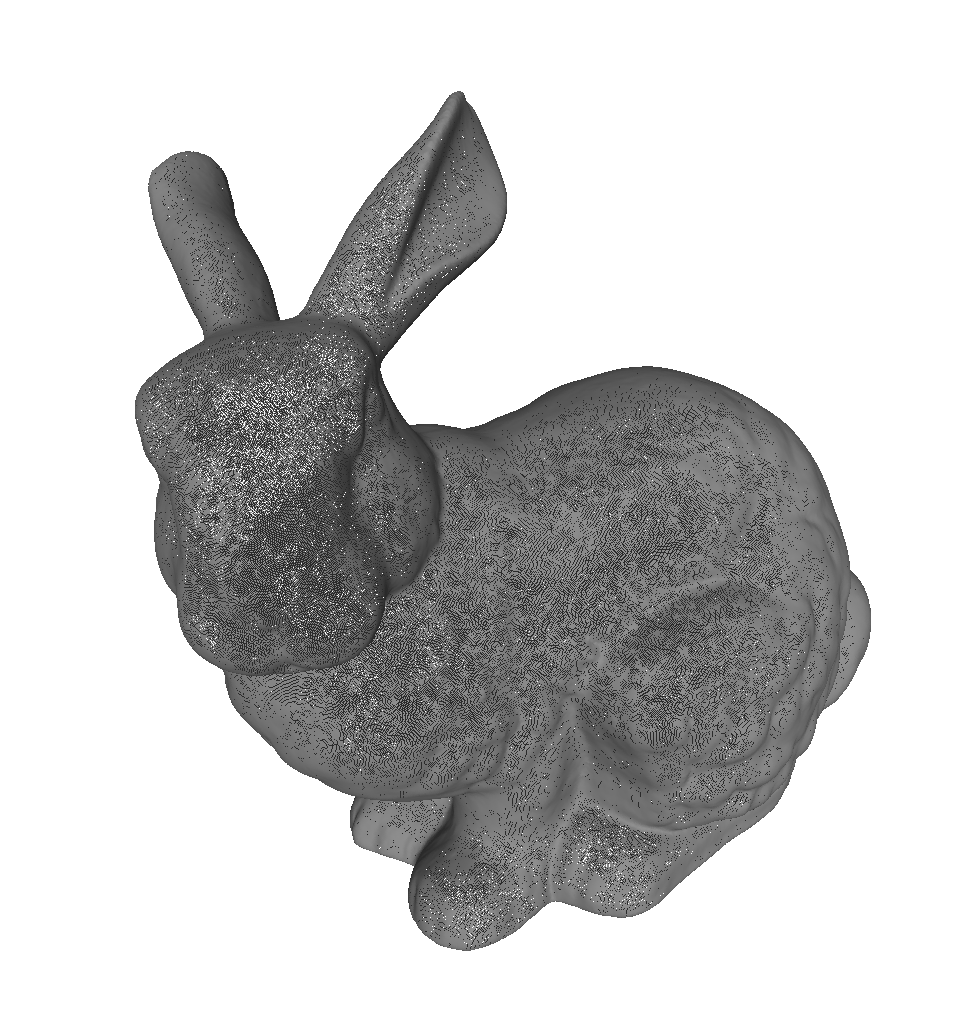}}
  \subfigure[]{\includegraphics[width=.49\linewidth]{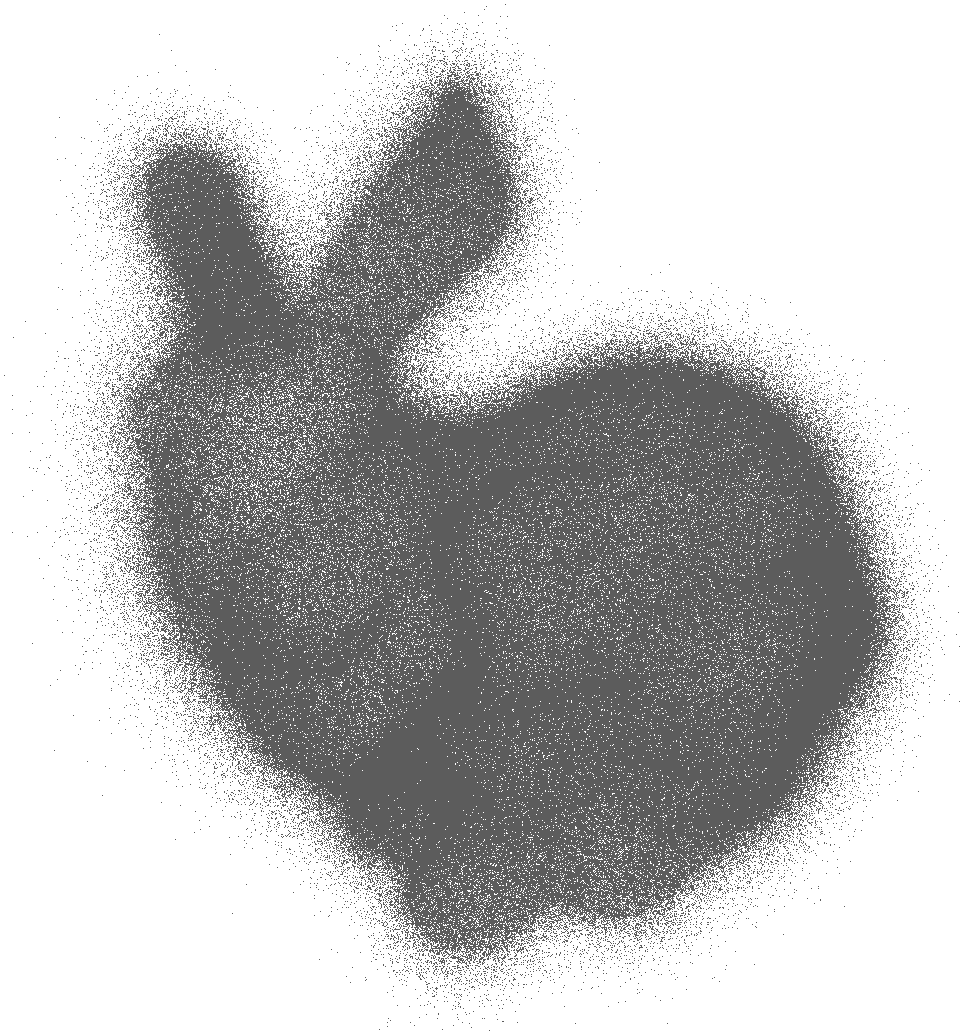}}
  \subfigure[]{\includegraphics[width=.49\linewidth]{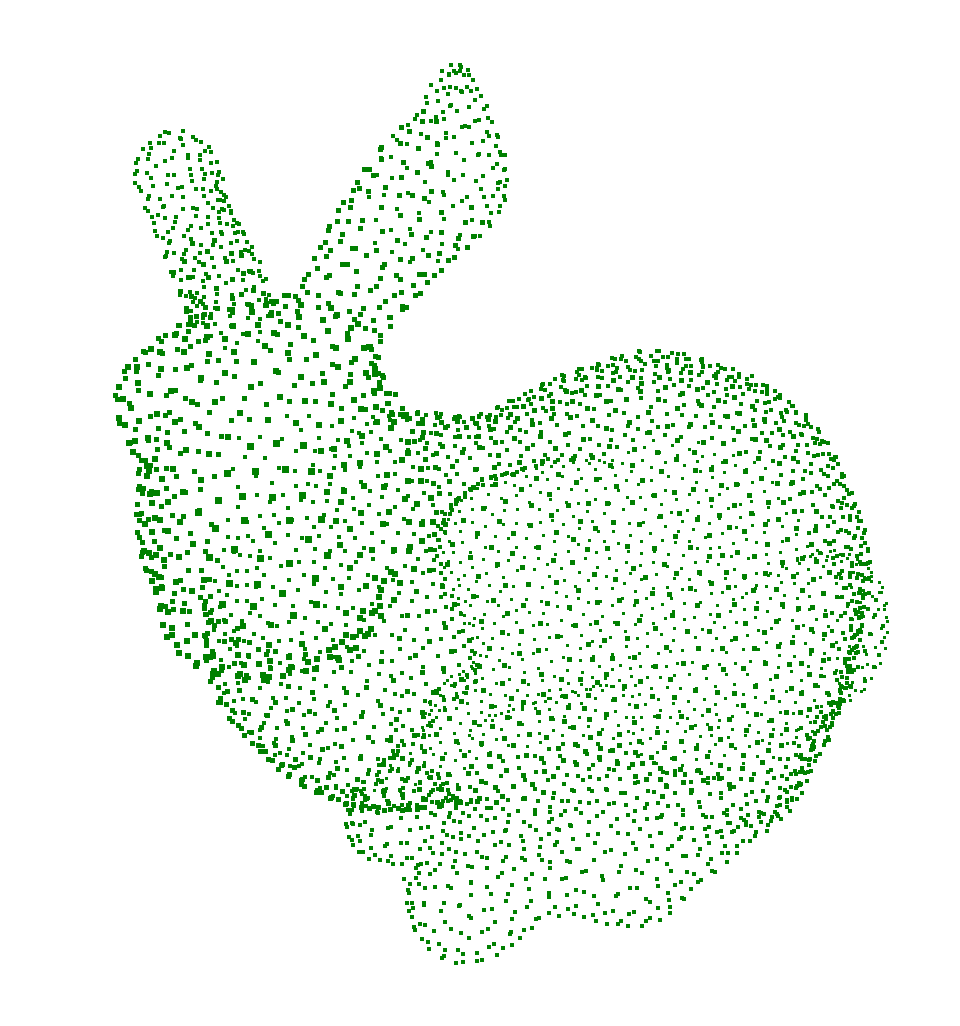}}
  \subfigure[]{\includegraphics[width=.49\linewidth]{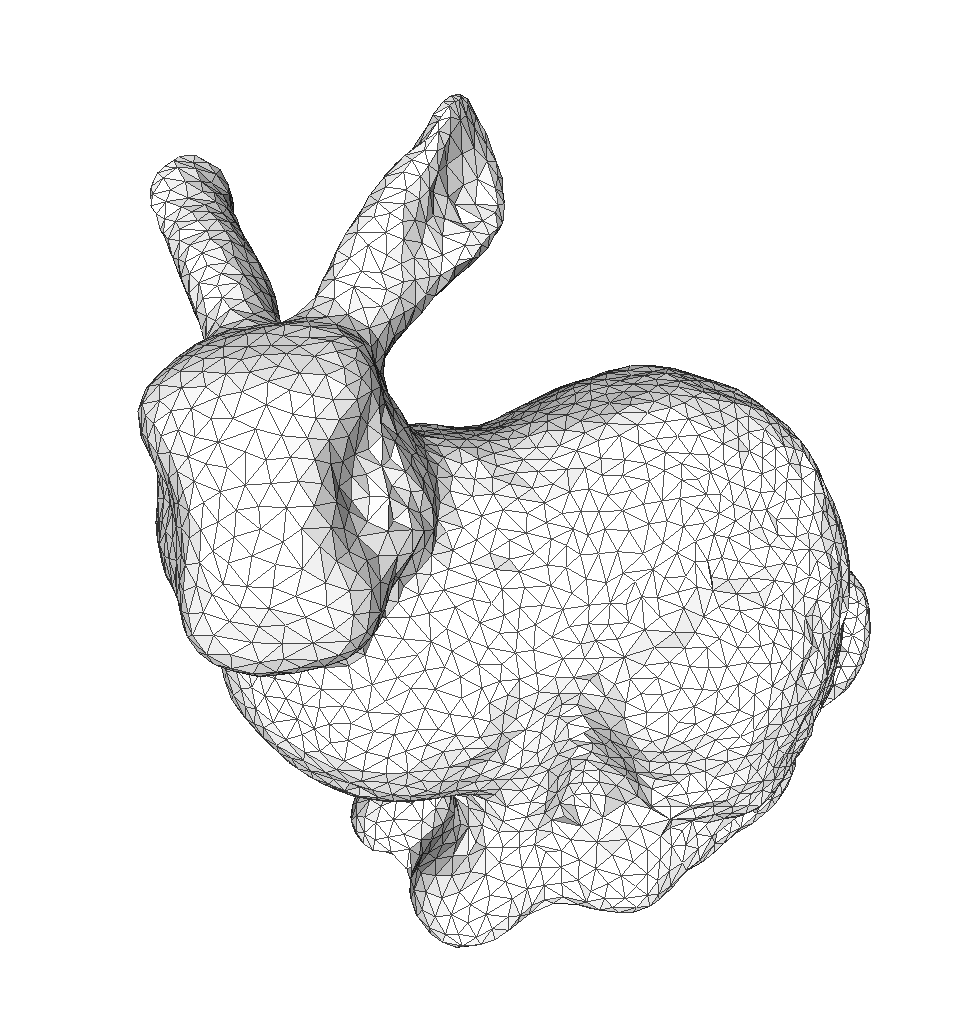}}
  \caption{\label{fig:seq-bunny}
A point sample (a) drawn with (almost) uniform probability from the surface of the Stanford bunny is then convolved (b) with isotropic Gaussian noise. Presented with the latter and with appropriate parameter settings (see text), the Neural Gas (c) attains a final configuration from which the original surface can be reconstructed faithfully (d).}
\end{figure}

In particular the focus of this work is providing an experimental assessment of NG behaviors like those exemplified in Fig.~\ref{fig:seq-bunny}, while giving some preliminary explanations. Although a deeper study is required, i.e. for finding better analytical models describing the NG behavior in these cases, the evidence presented suggests that this study can be feasible and worth performing.

\section{Methods}

\subsection{Neural Gas}

Given an input data distribution described by probability $P(\mathbf{v})$, where $\mathbf{v} \in \mathbb{R}^D$, a \emph{neural gas} (NG) \cite{Martinetz-etal93, Martinetz-Schulten94} is a set $\mathbf{W}$ of $k$ \emph{units}, each associated to a reference vector in $\mathbb{R}^D$:
$$\mathbf{W} := \{\mathbf{w}_i\},\ \mathbf{w}_i \in \mathbb{R}^D,\ i \in \{1,\ldots,k\}.$$
The NG algorithm adapts progressively the reference vectors in $\mathbf{W}$ to the input probability $P$ by repeating the following iteration:
\begin{enumerate}
  \item receive one signal $\mathbf{v}$ distributed as $P(\mathbf{v})$;
  \item update the reference vectors in $\mathbf{W}$ (see below);
  \item return to step 1.
\end{enumerate}
In each iteration, the  reference vectors in $\mathbf{W}$ are updated by
\begin{equation}
\label{eq:ng_update}
\Delta \mathbf{w}_{i} = \varepsilon\cdot h_{\lambda}(k_i(\mathbf{v}))(\mathbf{v} - \mathbf{w}_{i}),
\end{equation}
where $k_i(\mathbf{v}) := \#\{\mathbf{w}_{j} : \| \mathbf{v} - \mathbf{w}_{j} \| < \| \mathbf{v} - \mathbf{w}_{i} \|\}$ ($\#$ denotes the cardinality), $\varepsilon > 0$ is a real parameter, 
$$h_{0}(n) := \delta_{0n} \quad\text{and}\quad h_{\lambda}(n) := e^{-\frac{n}{\lambda}}, \text{ for } \lambda > 0$$
(throughout this paper, $\delta$ denotes the usual Kronecker delta function).

\subsubsection{Convergence}

It is proven in \cite{Martinetz-etal93} that the NG algorithm performs a \emph{stochastic gradient descent} (SGD) over the energy function:
\begin{equation}
\label{eq:ng_energy}
E_{NG}(\mathbf{W}) = \frac{1}{2C_{\lambda}}\sum_{i=1}^k{\int_V{P(\mathbf{v}) h_{\lambda}(k_i(\mathbf{v})) (\mathbf{v} - \mathbf{w}_{i})^2 d\mathbf{v}}}
\end{equation}
where $V$ is the support of probability $P(\mathbf{v})$ and
$$C_{\lambda} := \sum_{i=0}^{k-1}{h_{\lambda}(i)}.$$
In particular, in \cite{Martinetz-etal93} it is proved that
$$\frac{\partial E_{NG}}{\partial \mathbf{w}_{i}} = - \frac{1}{C_{\lambda}}\int_V{P(\mathbf{v}) h_{\lambda}(k_i(\mathbf{v})) (\mathbf{v} - \mathbf{w}_{i}) d\mathbf{v}},$$
which makes equation \eqref{eq:ng_update} an SGD update law. In keeping with this, an NG can be made to converge to a steady configuration by choosing values of $\varepsilon$ that decrease exponentially with the iterations of the algorithm:
\begin{equation}
\label{eq:decaying_epsilon}
  \varepsilon(t) := \varepsilon_i\left(\frac{\varepsilon_f}{\varepsilon_i}\right)^{t/T},
\end{equation}
where $\varepsilon_i$ and $\varepsilon_f$ are the initial and final values, respectively, $T$ is the total number of iterations and $t$ is the current iteration.

\subsubsection{Relation with K-means}

When $\lambda \rightarrow 0$, equation \eqref{eq:ng_update} becomes equivalent to
$$\Delta \mathbf{w}_{i} = \varepsilon\cdot \delta_{ik(\mathbf{v})} \cdot (\mathbf{v} - \mathbf{w}_{i}),$$
where $k(\mathbf{v})$ is the function that returns the index of the closest neighbor in $\mathbf{W}$ to the input signal $\mathbf{v}$. This is the update law of the well-known algorithm named K-means \cite{lloyd1982least,macqueen1967some}, which performs an SGD over the \emph{distortion error}
$$E_{K}(\mathbf{W}) = \frac{1}{2}\int_V{P(\mathbf{v}) (\mathbf{v} - \mathbf{w}_{k(\mathbf{v})})^2 d\mathbf{v}}.$$

\subsubsection{Magnification}
\label{sec:magnification}

Considering the average position update of a unit with respect to $P$
$$\langle \Delta \mathbf{w}_{i} \rangle = \varepsilon \cdot \int_V{P(\mathbf{v}) h_{\lambda}(k_i(\mathbf{v})) (\mathbf{v} - \mathbf{w}_{i}) d\mathbf{v}}$$
under specific assumptions to be discussed below, in \cite{Martinetz-etal93} it is also proven that: 
\begin{equation}
\label{eq:ng_avg_update}
\langle \Delta \mathbf{w}_{i} \rangle \propto \frac{1}{\rho(\mathbf{w}_{i})^{\frac{1}{\alpha}}}\left((\partial_{\mathbf{v}}P)(\mathbf{w}_{i}) - \frac{1}{\alpha} P(\mathbf{w}_{i})\frac{(\partial_{\mathbf{v}}\rho)(\mathbf{w}_{i})}{\rho(\mathbf{w}_{i})}\right)
\end{equation}
where 
$$\alpha := \frac{d}{d+2},$$
$\rho(\mathbf{w}_{i})$ is the density of units in $\mathbf{W}$ at $\mathbf{w}_{i}$, $\partial_{\mathbf{v}}\rho$ and $\partial_{\mathbf{v}}P$ are the gradients with respect to $\mathbf{v} \in V$ of, respectively, the density $\rho$ and the probability $P$, and $d$ is the dimension of the support $V$. The solution at equilibrium, i.e. when $\langle \Delta \mathbf{w}_{i} \rangle = 0$, entails a power law:
\begin{equation}
\label{eq:ng_magnification}
\rho(\mathbf{w}_{i}) \propto P(\mathbf{w}_{i})^{\alpha}.
\end{equation}
In many related works (see for instance \cite{bishop1997magnification, jain2004forbidden, claussen2005magnification, villmann2006magnification, hammer2007magnification}) this particular power law is called \emph{magnification}, while $\alpha$ is deemed the \emph{magnification factor}.

From a technical standpoint, the derivation of \eqref{eq:ng_magnification} relies on three main assumptions:
\begin{itemize}
  \item[A.1:] $\rho$ is analytic in $\mathbb{R}^D$
  \item[A.2:] $P$ is analytic in $\mathbb{R}^D$
  \item[A.3:] $\lambda > 0$ and small with respect to the \emph{curvature} of $\rho$ and $P$.
\end{itemize}
Assumption A.1 means considering the discrete set $\mathbf{W}$ in the limit of a continuous density while A.2 is in often true `per se', with typical input probabilities. Altogether, A.1 and A.2 allow expanding $\rho$ and $P$ into Taylor series, whereas assumption A.3 allows considering only the first terms in these. From another point of view, A.3 entails `flatness' of $P$, at least in the scale of $\lambda$, hence ruling out higher order contributions from its geometry.  

\subsection{Numerical Experiments}

\subsubsection{Estimating Probability and Density}
The estimation of $P(\mathbf{w}_{i})$ and $\rho(\mathbf{w}_{i})$ is a crucial aspect for interpreting the results of numerical experiments and therefore requires an appropriate technique. In other works, like \cite{merenyi2007explicit}, the ambient space is divided into bins of suitable size, while  in \cite{hammer2007magnification} a Parzen window estimator is used. The estimation technique adopted here is described in \cite{biau2011weighted}. It is based on a weighted $k$-distance defined in \cite{chazal2011geometric} and that, in the discrete case, takes the form
\begin{equation}
\label{eq:mu_distance}
d_{C,m_0}(\mathbf{x}) := \left(\frac{1}{m_0} \sum_{i=1}^{m_0}{\|\mathbf{x} - \mathbf{c}_i(\mathbf{x})\|^2} \right)^{\frac{1}{2}},
\end{equation}
where $C$ is a finite point cloud in $\mathbb{R}^D$, $m_0 \in \mathbb{N}^+$ is a parameter, and $\mathbf{c}_i(\mathbf{x})$ is the $i$-th nearest neighbor to $\mathbf{x}$  in $C$. In the case discussed here, $C$ can either represent an empirical sample of points drawn from $P$ or the set $\mathbf{W}$ of NG units. In \cite{chazal2011geometric} it is shown that \eqref{eq:mu_distance} is \emph{distance-like}, i.e. it shares many properties of a distance function, and that its level sets preserve the geometrical and topological properties of the corresponding level sets of the probability $P$ from which the point cloud has been drawn. Based on \eqref{eq:mu_distance}, the estimator proposed in \cite{biau2011weighted} is
\begin{equation}
\label{eq:p_estimator}
\hat{P}_{C,m_0}(\mathbf{x}) := \frac{1}{N V_D} \left(\frac{\sum_{i=1}^{m_0}{i^{\frac{2}{D}}}}{m_0\;d_{C,m_0}^2(\mathbf{x})}\right)^{\frac{D}{2}},
\end{equation}
where $N$ is the number of points in $C$ and $V_D$ is the volume of the $D$-dimensional unit ball. Reportedly, the estimator \eqref{eq:p_estimator} is optimal for
$$m_0 \simeq N^{\frac{4}{D+4}}.$$
In this work \eqref{eq:p_estimator} has been used for estimating $P(\mathbf{w}_i)$ and $\rho(\mathbf{w}_i)$ respectively; the latter, in fact, can be normalized into the probability
$$R(\mathbf{w}_i) := \frac{\rho(\mathbf{w}_i)}{k}.$$
For simplicity, however, in this work we will continue to refer to the unnormalized density $\rho(\mathbf{w}_i)$.

\subsubsection{Entropy}
Given a final NG configuration, the probability of a generic unit $i$ of being `a winner' in $\mathbf{W}$, i.e. of being the nearest neighbor to a generic input signal $\mathbf{v}$ having probability $P(\mathbf{v})$, can be estimated by
$$p_i = \frac{s_i}{s},$$
where $s$ is the total number of input signals and $s_i$ is the number of input signals in the Voronoi cell of unit $i$ or, in other words, is the number of input signals having $\mathbf{w_i}$ as their closest neighbor in $\mathbf{W}$.

The \emph{Shannon entropy} function is defined as
\begin{equation}
\label{eq:entropy}
H := -\sum_{i=1}^k{p_i \ln p_i}.
\end{equation}
The maximum of $H$ is attained when all $k$ units have equal probability of being the nearest neighbor of a random input signal sampled from $P$. In that case, the value of Shannon entropy is
$$H = -\sum_{i=1}^k{\frac{1}{k} \ln \frac{1}{k}} = - \ln \frac{1}{k} = \ln k.$$

\subsubsection{Implementation and execution}
The software for the experiments has been written in Java; all experiments have been performed with a 64-bit Java virtual machine running on an Intel$^\circledR$ Xeon$^\circledR$ CPU E-1240 v3 @ 3.40 GHz computer with 8GB of RAM. The Meshlab software tool \cite{cignoni2008meshlab} has been used for 2D and 3D processing of point clouds and for the geometrical and topological validation of results.

\section{Results}

In the experiments decribed here, the initial positions of NG units in $\mathbf{W}$ have been randomly chosen with probability $P$. For this reason, the only parameters in \eqref{eq:ng_update} that determine the NG behaviour are $\varepsilon, \lambda$ and $k$.

In this study, the decaying law \eqref{eq:decaying_epsilon} has been adopted  for parameter $\varepsilon$, with $\varepsilon_i=0.1$ and $\varepsilon_f=0.0001$, to obtain steady NG configurations that make it easier the assessment of relevant properties. The combined effect of the remaining parameters $\lambda$ and $k$, with different input data distributions, has been studied in more than 3500 experiments, each intended as a complete run of the NG algorithm. 

Due to the general field of interest of this work, only 2D and 3D ambient spaces have been considered. All input data distributions for the experiments were generated from probabilities in the general form of the convolution described:
$$\mathcal{U}(\text{\emph{Shape}}) \ast N(\mathbf{0},\mathbf{\theta}).$$
Fundamental \emph{shapes} like individual points, the unit circle $\mathbb{S}^1$, the unit disk $\mathbb{D}$, the unit sphere  $\mathbb{S}^2$ and the unit ball  $B(\mathbf{0},1)$ have been considered. As an example of a more complex shape, the Stanford bunny\footnote{This point cloud, together with other popular examples, can be downloaded at \url{http://graphics.stanford.edu/data/3Dscanrep/ }}, which is a classical benchmark in the field of geometrical shape reconstruction, has been also adopted. For the noise model to be convolved, the class of multivariate normal distributions
$\mathcal{N}(\mathbf{0},\mathbf{\Sigma)}$ with isotropic covariance matrix, i.e.
\begin{equation}
  \label{eq:covariance}
  \mathbf{\Sigma} := \sigma^2 \mathbf{I}.
\end{equation}
has been considered as the first option.

Two other noise models have been considered. One is a sinusoidal probability with density
\begin{equation}
  \label{eq:sinusoidal}
  N(\mathbf{v};\mathbf{0}, r) := \frac{\pi}{2} \sin\left( \frac{\pi}{2} + \frac{\pi}{2}\frac{\|\mathbf{v}-\mathbf{0}\|}{r}\right)
\end{equation}
whenever $\|\mathbf{v} - \mathbf{0}\| < r$ and $0$ elsewhere in $\mathbb{R}^D$. Another is the uniform density over a ball of given radius $r$:
$$\mathcal{U}(B(\mathbf{0},r)).$$
To ensure the possibility of estimating probability densities at the points attained by NG units in the final configurations, all data distributions were generated in advance, as standalone point clouds, and then presented to the NG algorithm.

For reasons of space and readability, in the rest of this section, the results obtained will not be described systematically but rather by following a logical pathway that illustrates the most relevant findings through the more relevant steps.

\begin{figure}[h]
  \centering
  \subfigure[]{\includegraphics[width=.34\linewidth]{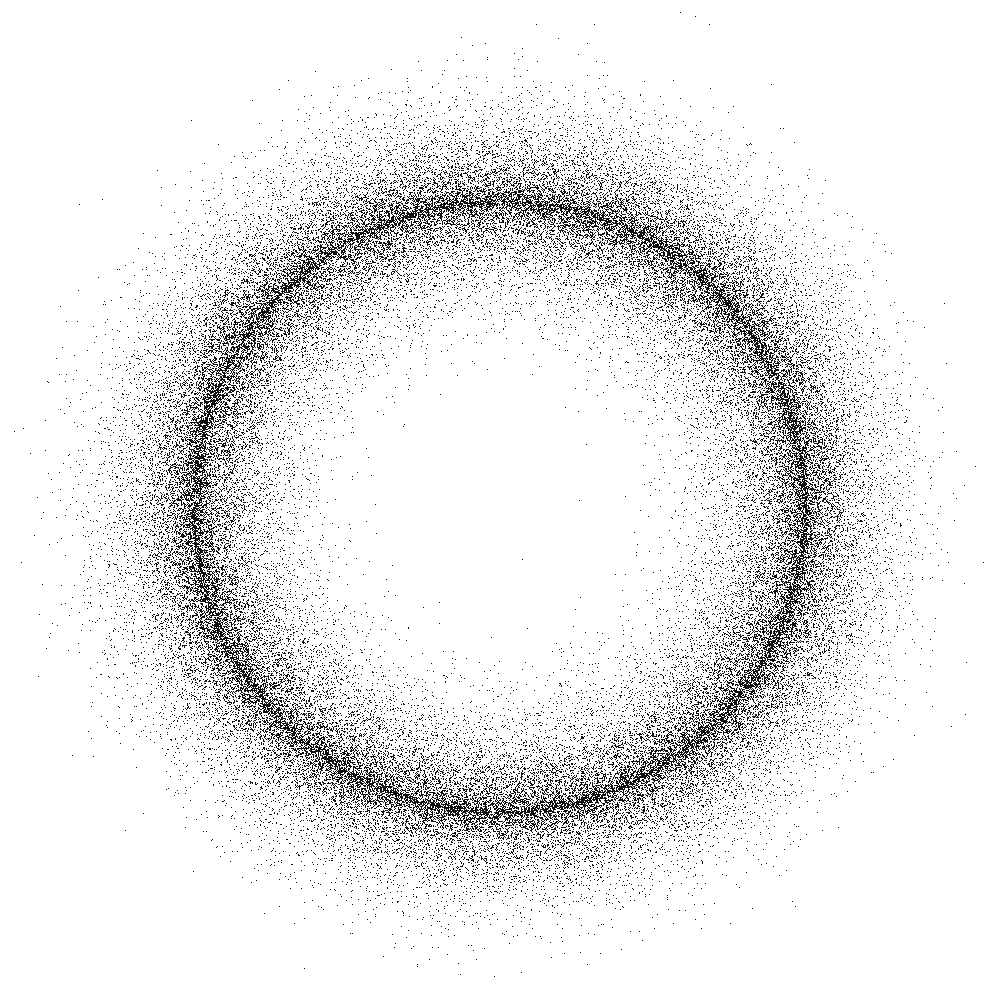}}
  \hspace{.05\linewidth}
  \subfigure[]{\includegraphics[width=.34\linewidth]{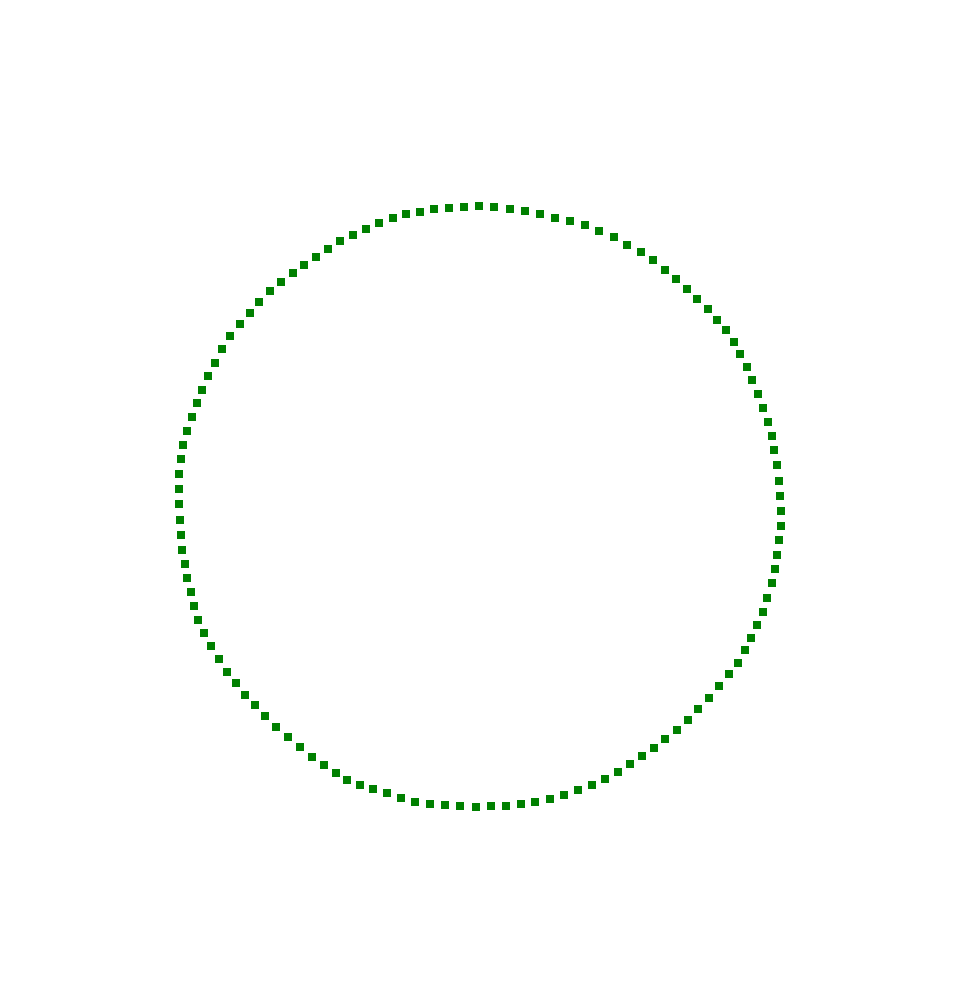}}
  \subfigure[]{\includegraphics[width=.34\linewidth]{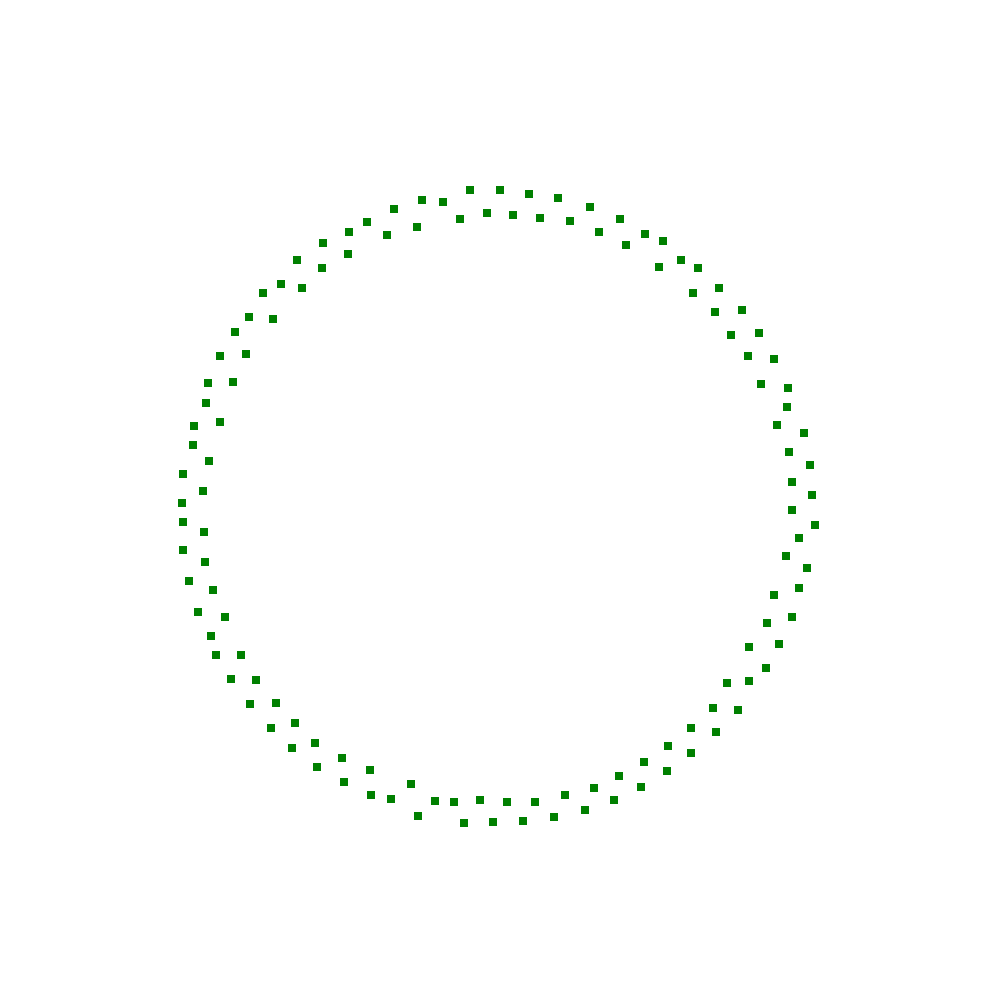}}
  \hspace{.05\linewidth}
  \subfigure[]{\includegraphics[width=.34\linewidth]{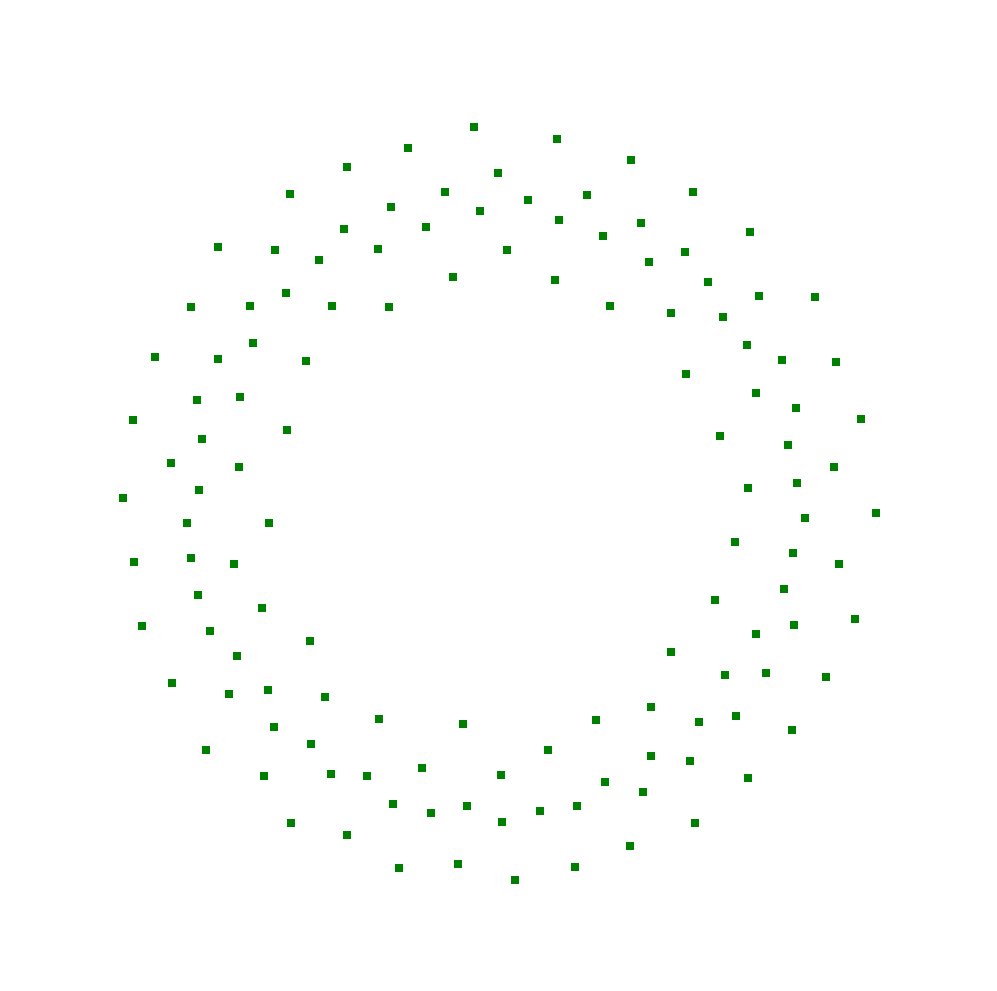}}
  \caption{\label{fig:decaying_lambda}
An experiment with an NG with decaying $\lambda$ values ($\lambda_i=8.0$, $\lambda_f=0.05$, $T=4\text{M}$) presented with a 2D uniform sample from the unit circle convolved with normal noise (a). After 150K iterations, the NG (b) is very close to the unit circle then, after 500K iterations (c), as $\lambda$ decreases, the NG begins dispersing until it reaches its final configuration (d), after 4M iterations.}
\end{figure}

\subsection{Decaying lambda?}

Assumption A.3 about the $\lambda$ parameter (see section \ref{sec:magnification}) does not translate immediately into numerical constraints and is therefore difficult to enforce in practical experiments. Partly for this reason, it is suggested in \cite{Martinetz-etal93} to make $\lambda$ decay exponentially as the algorithm progresses, with the same law adopted for $\varepsilon$:
\begin{equation}
  \label{eq:decaying_lambda}
  \lambda(t) := \lambda_i\left(\frac{\lambda_f}{\lambda_i}\right)^{t/T}.
\end{equation}
Note however that this decay law is not part of the model described by \eqref{eq:ng_avg_update}. In passing, the same decay law is also suggested in Kohonen's \emph{Self-Organizing Map} (SOM) algorithm since `it produces an improvement in the global order of units' \cite{kohonen1990self}.

Fig.~\ref{fig:decaying_lambda} shows the typical behavior of an NG with a decaying $\lambda$. In this experiment the 2D-embedded input probability is
$$\mathcal{U}(\mathbb{S}^1) \ast \mathcal{N}(\mathbf{0},\mathbf{\Sigma)}$$ 
where $\mathbb{S}^1$ is the unit circle and the $\mathcal{N}$ is 2D isotropic multivariate normal, with $\sigma := 0.18$ (see \eqref{eq:covariance}). Initially, when the value of $\lambda$ is large, the NG assumes a configuration that is closer to $\mathbb{S}^1$; then, as the algorithm progresses and $\lambda$ decreases, the NG attains a more dispersed configuration that resembles closely the input distribution. As it can be seen, during the process, the NG undergoes substantial changes of `shape' which are not necessarily reflected in the final configuration.

Given the objectives of this work, these intermediate NG configurations are indeed relevant and therfore we focused on experiments with \emph{constant} $\lambda$ values.
\begin{figure}[h]
  \centering
  \subfigure[]{\includegraphics[width=.42\linewidth]{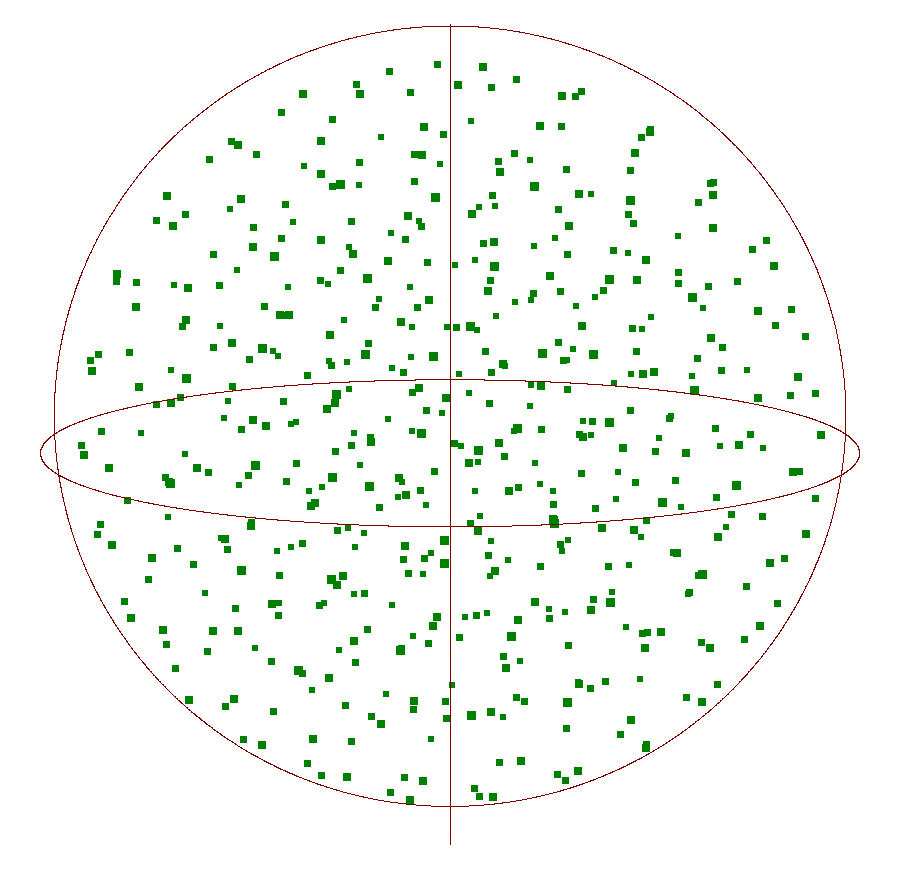}}
  \hspace{.05\linewidth}
  \subfigure[]{\includegraphics[width=.42\linewidth]{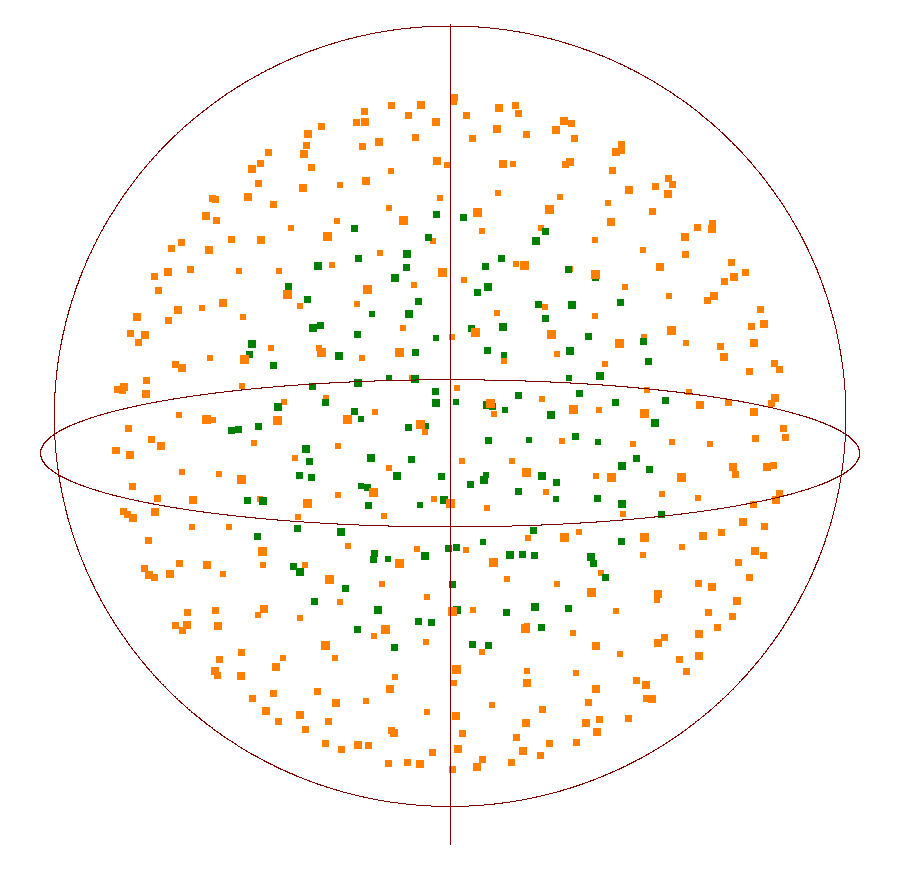}}
  \subfigure[]{\includegraphics[width=.42\linewidth]{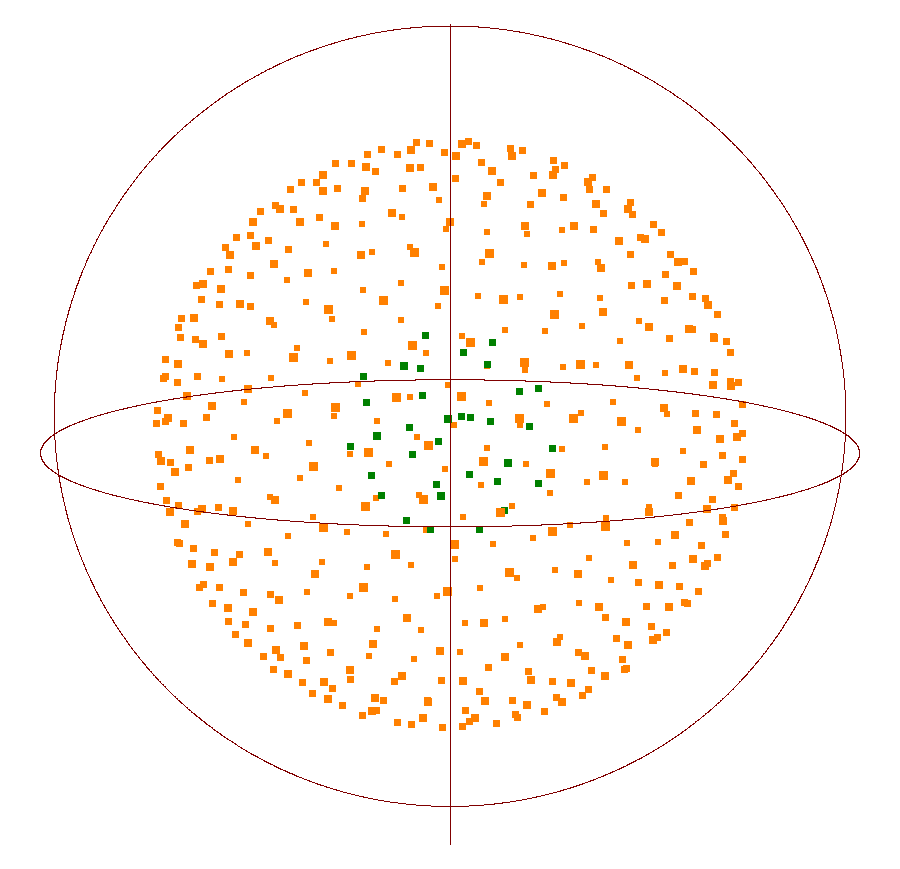}}
  \hspace{.05\linewidth}
  \subfigure[]{\includegraphics[width=.42\linewidth]{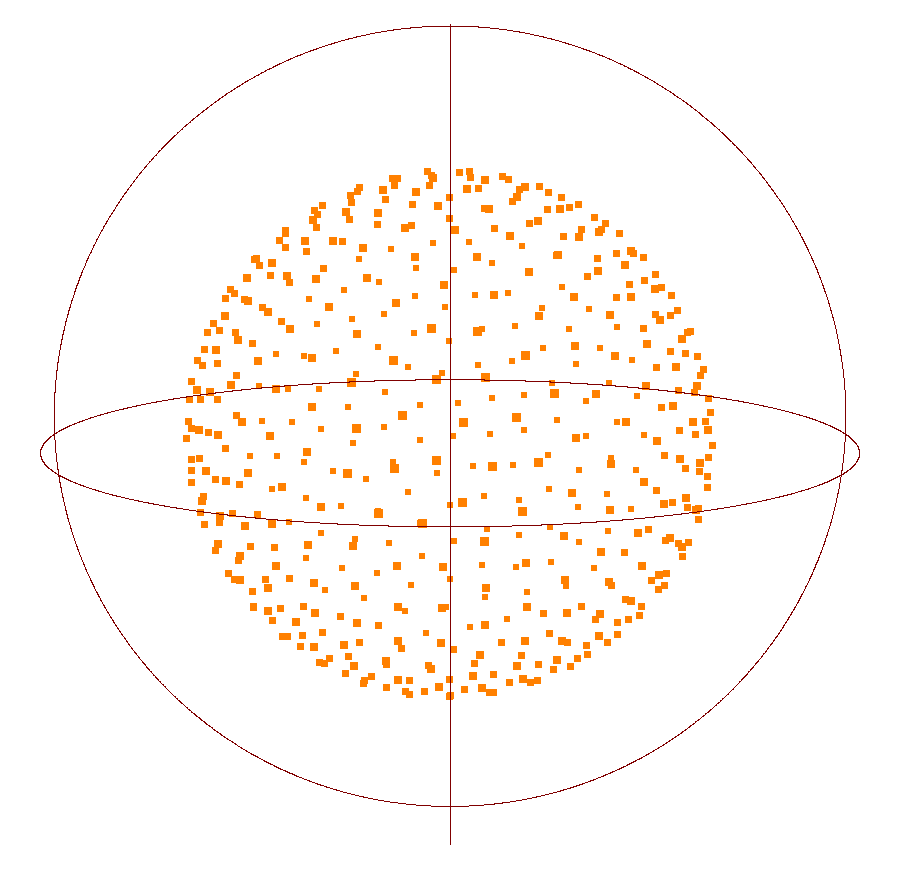}}
  \caption{\label{fig:uniform_ball}
Four final NG configurations for the same uniform probability over a 3D unit ball (shown by the external `trackball'), with $k=512$ and different (constant) $\lambda$ values. For $\lambda = 0$ (a) the NG distributes uniformly as well. With greater values of $\lambda$, an outer sphere (in orange) separates from an inner ball (in dark green), as shown for $\lambda=4$ (b) and $\lambda=16$ (c). For $\lambda \ge 32$ (d) the NG disposes itself on the surface of an empty sphere.}
\end{figure}

\subsection{Shape is relevant}

The changes in NG configurations and the relevance of geometrical features in the input probability $P$ are shown by the experiments in Fig.~\ref{fig:uniform_ball}, which describes four final NG configurations, each corresponding to different values of $\lambda$, for the same 3D-embedded input probability 
\begin{equation}
  \label{eq:p_uniform}
  \mathcal{U}(B(\mathbf{0},1))
\end{equation}
 where $\mathcal{U}$ is the uniform distribution and $B(\mathbf{0},1)$ is the 3D unit ball centered in the origin.  

As it can be seen, depending on $\lambda$, the final NG configurations can be grouped into three main classes, or `phases', defined in terms of the distinct describing topologies (ball; sphere with an inside ball; empty sphere)\footnote{The notion of topology-based `phase' transitions for NG is left intuitive here for ease of discussion. A more precise definition, however, can be given along the lines described in \cite{chazal2009sampling}.}. 

In addition, further experiments show that these transitions also depend on $k$ and occur with remarkable regularity; in all experiments performed with the input probability above, with integer $\lambda$ values ranging from 0 to 32 and $k$ equal, respectively, to  $32,64,128,256,512,1024$ and $2048$, the transition between the topology in Fig.~\ref{fig:uniform_ball}(c) and that in Fig.~\ref{fig:uniform_ball}(d) always occurs at $\lambda = \frac{k}{16}$.

\begin{figure}[h]
  \centering
  \subfigure[]{\includegraphics[width=.34\linewidth]{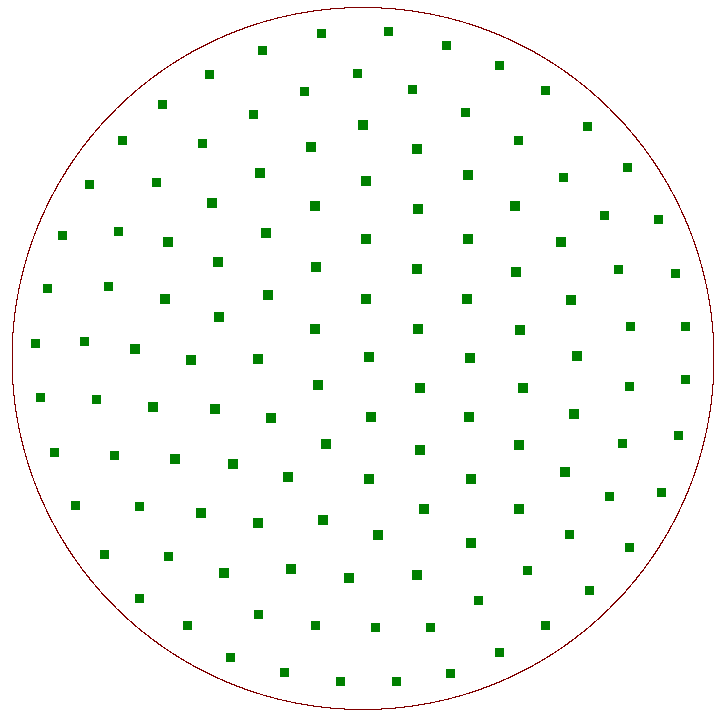}}
  \hspace{.05\linewidth}
  \subfigure[]{\includegraphics[width=.34\linewidth]{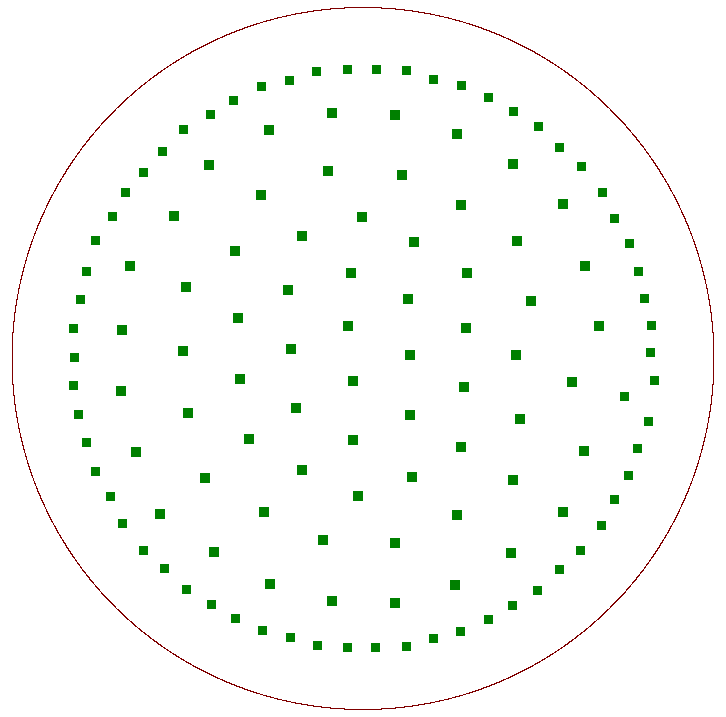}}
  \subfigure[]{\includegraphics[width=.34\linewidth]{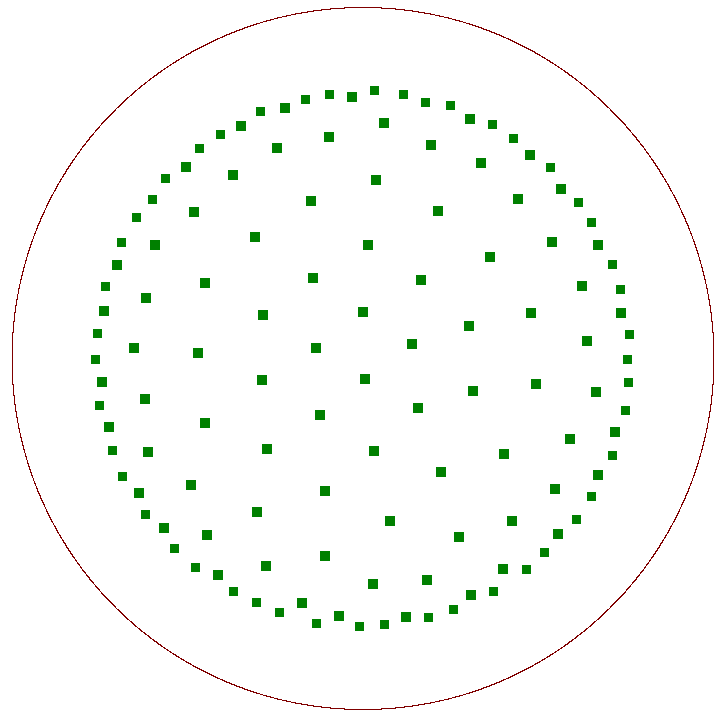}}
  \hspace{.05\linewidth}
  \subfigure[]{\includegraphics[width=.34\linewidth]{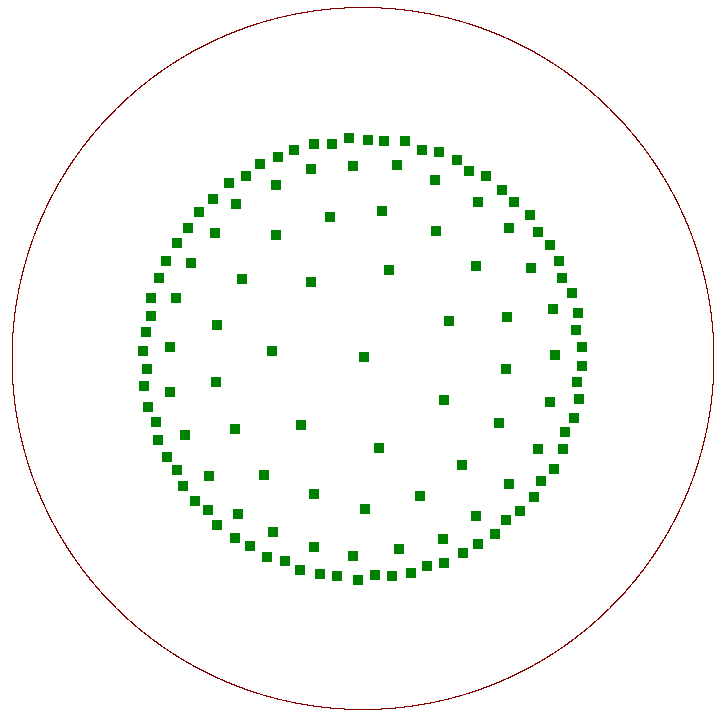}}
  \caption{\label{fig:uniform_disk}
A 2D uniform probability over a unit disk (shown by the outer circle) does not induce an NG a behavior similar to that in Fig.~\ref{fig:uniform_ball}, i.e with a uniformly-sampled 3D ball as input. The final NG configurations represented here are for $k=128$ and $\lambda$ equal to $0$, $4$, $8$ and $16$ respectively.}
\end{figure}

This 3D phenomenon is much less evident in 2D, as described by Fig.~\ref{fig:uniform_disk}. In the experiments performed, in the final NG configurations with a 2D-embedded probability of  $\mathcal{U}(\mathbb{D})$, where $\mathbb{D}$ is the unit disk, the changes of the describing topology are much less dramatic than in the 3D case.

\begin{figure}[h]
  \centering
  \subfigure[]{\includegraphics[trim=3 4 3 3,clip,width=.6\linewidth]{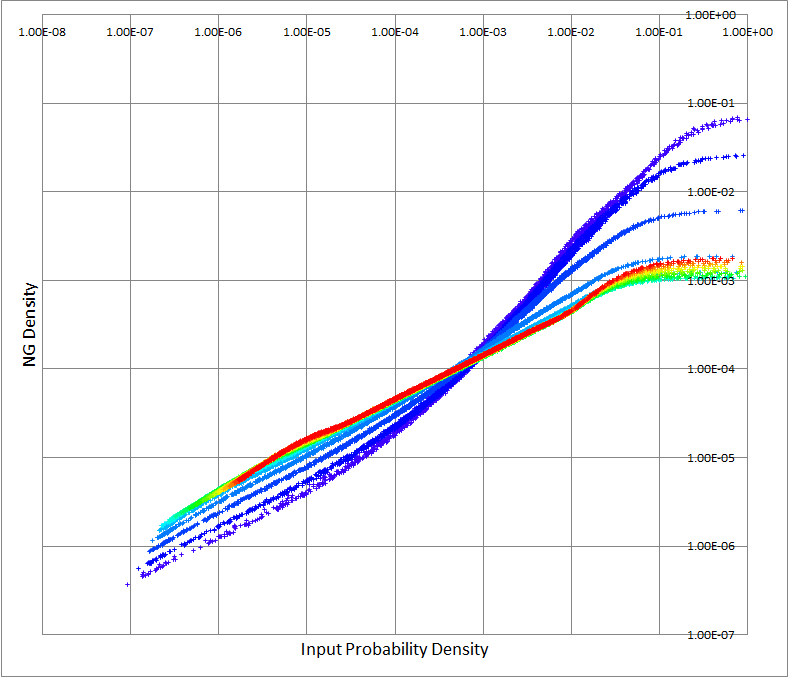}}
  \subfigure[]{\includegraphics[trim=3 4 3 3,clip,width=.6\linewidth]{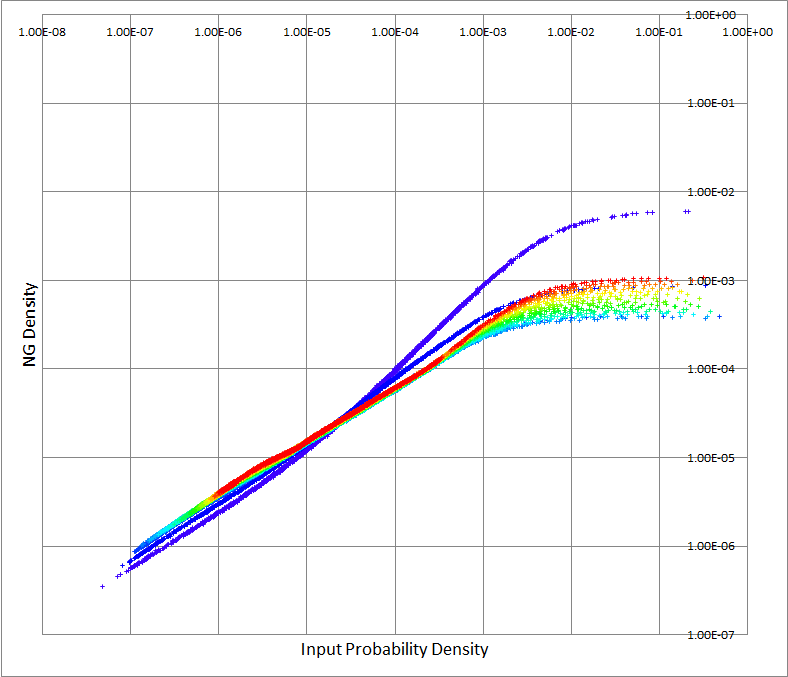}}
  \caption{\label{fig:gaussian_power_law}
Log-log density plots of final NG configurations with, respectively, 2D (a) and 3D (b) multivariate, isotropic normal probabilities. with constant $\lambda$ values ranging from $0$ to $18$ (shown with increasingly warm colors). }
\end{figure}

\subsection{Magnification}

Further experiments have been performed with the noise model alone, i.e. without a \emph{shape}, using 2D and 3D multivariate normal probabilities $\mathcal{N}(\mathbf{0},\mathbf{\Sigma})$ with isotropic covariance matrix. The diagrams in Fig.~\ref{fig:gaussian_power_law} compare on a log-log plot the densities $\rho(\mathbf{w_i})$ attained in final NG configurations and the probabilities $P(\mathbf{w_i})$ above, both estimated with \eqref{eq:p_estimator} and the suggested optimal value of $m_0$.

Different data series represented  in the diagrams correspond to $\lambda$ values ranging from $0$ to $18$ (data series with higher $\lambda$ values are shown in warmer colors). All of these have been obtained with $k=2048$. As shown in the log-log plots, as  $\lambda$ grows larger the power law becomes evident. More precisely, for $\lambda \ge 5$ there exists a limit upper density to which the densest $5\%$ NG units tend asymptotically, while the remaining $95\%$ NG units follow the power law. In addition, the estimated values of $\alpha$ is in substantial agreement with the value predicted by \eqref{eq:ng_magnification}; e.g. for $\lambda=5$ the estimated $\alpha$ values for $d=2$ and $d=3$ are $0.5132$ and $0.5995$ respectively, versus predicted values of $0.5$ and $0.6$.

\begin{figure}[h]
  \centering
  \subfigure[]{\includegraphics[trim=3 4 3 3,clip,width=.6\linewidth]{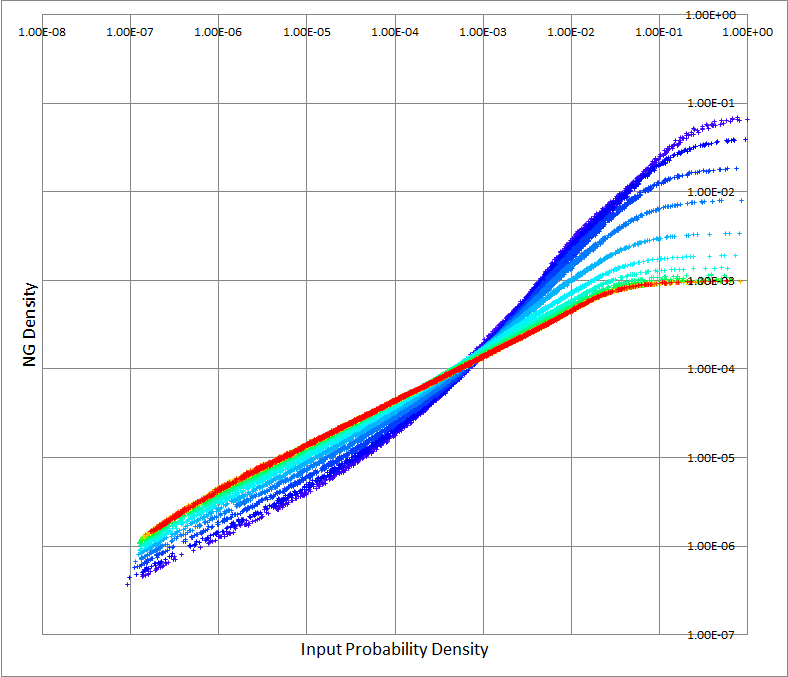}}
  \subfigure[]{\includegraphics[trim=3 4 3 3,clip,width=.6\linewidth]{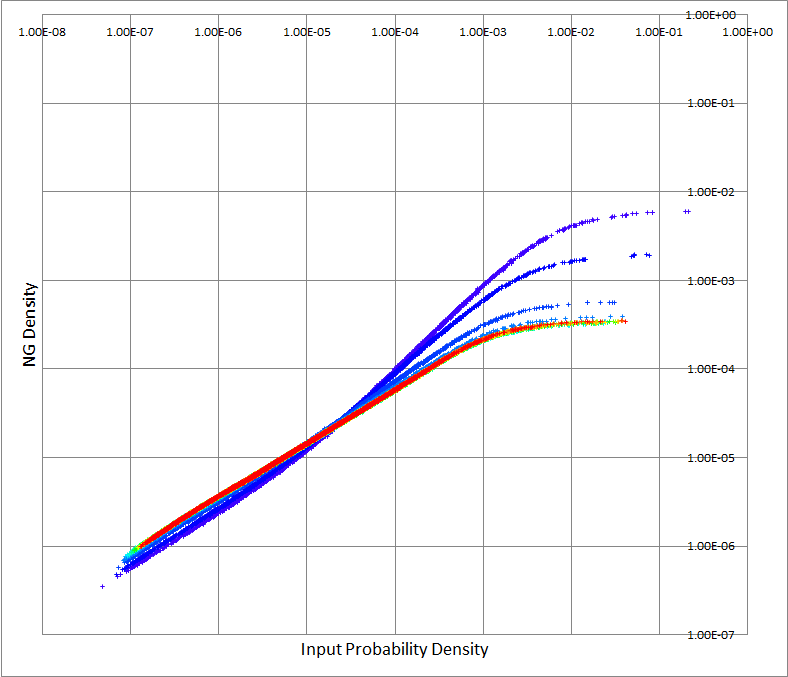}}
  \caption{\label{fig:gaussian_power_law_decaying}
Log-log density plots for the same experimental settings as in Fig.~\ref{fig:gaussian_power_law} but with decaying $\lambda$ values (see text).}
\end{figure}

In passing, Fig.~\ref{fig:gaussian_power_law_decaying} shows the log-log plots obtained in the same conditions as for the experiemnts in Fig.~\ref{fig:gaussian_power_law} but with $\lambda$ values that decay following \eqref{eq:decaying_lambda} from initial $\lambda_i$ in the range 1--18 to $\lambda_f=0.001$ in 5M iterations. Although the log-log plots in Fig.~\ref{fig:gaussian_power_law_decaying} are not very different from those in Fig.~\ref{fig:gaussian_power_law}, it can be seen by comparing the two figures that when $\lambda$ is kept constant the region occupied by NG densities tends to shrink as $\lambda$ grows larger.

\begin{figure}[h]
  \centering
  \subfigure[]{\includegraphics[width=.45\linewidth]{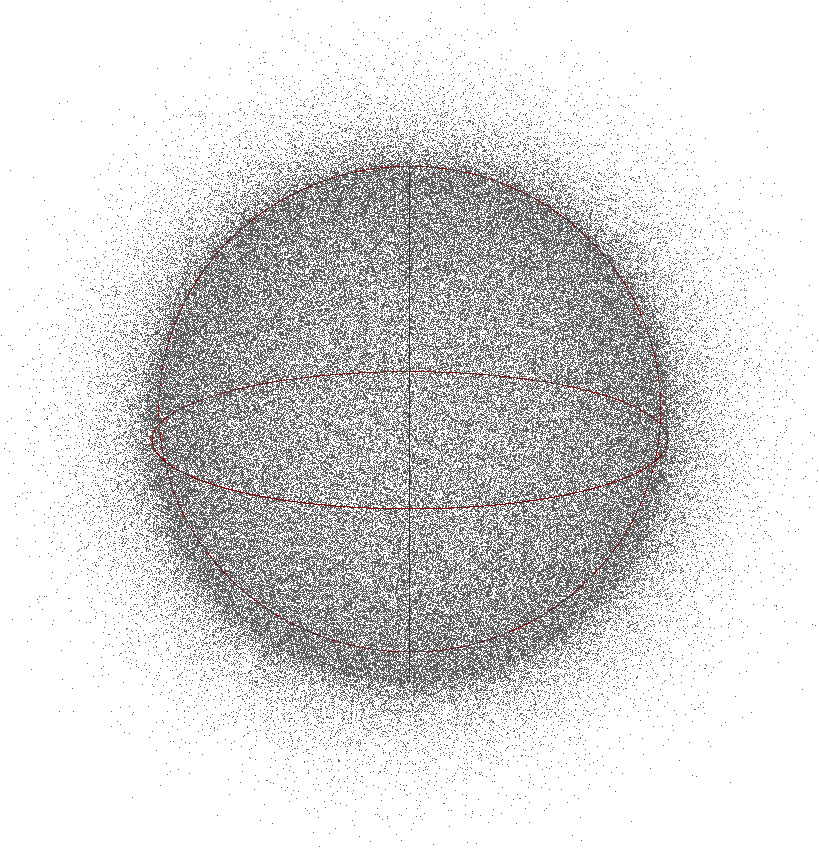}}
  \subfigure[]{\includegraphics[width=.45\linewidth]{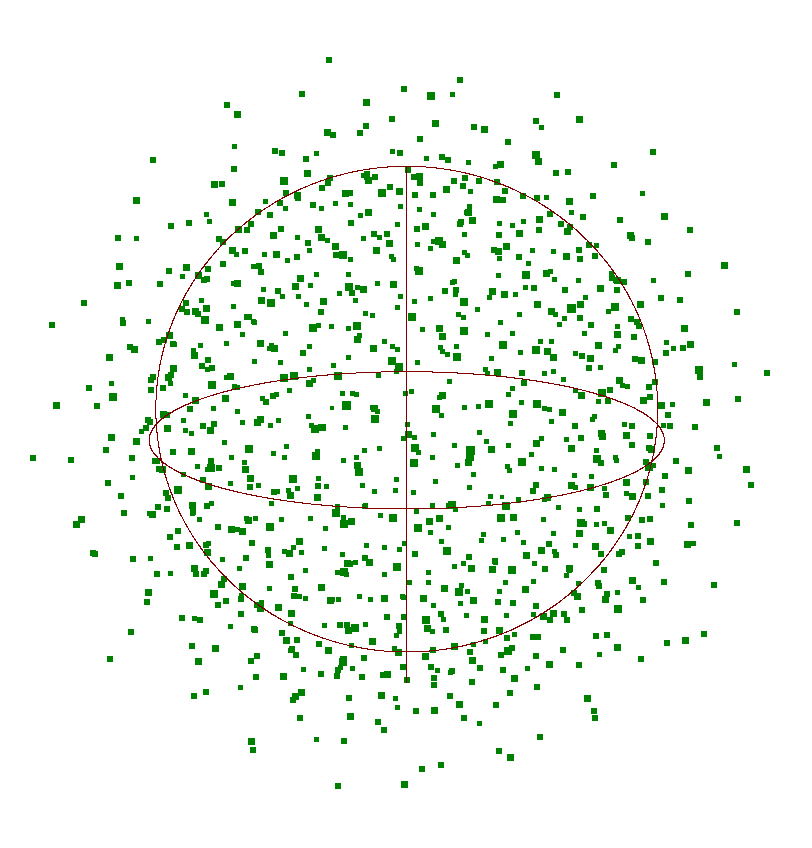}}
  \subfigure[]{\includegraphics[width=.45\linewidth]{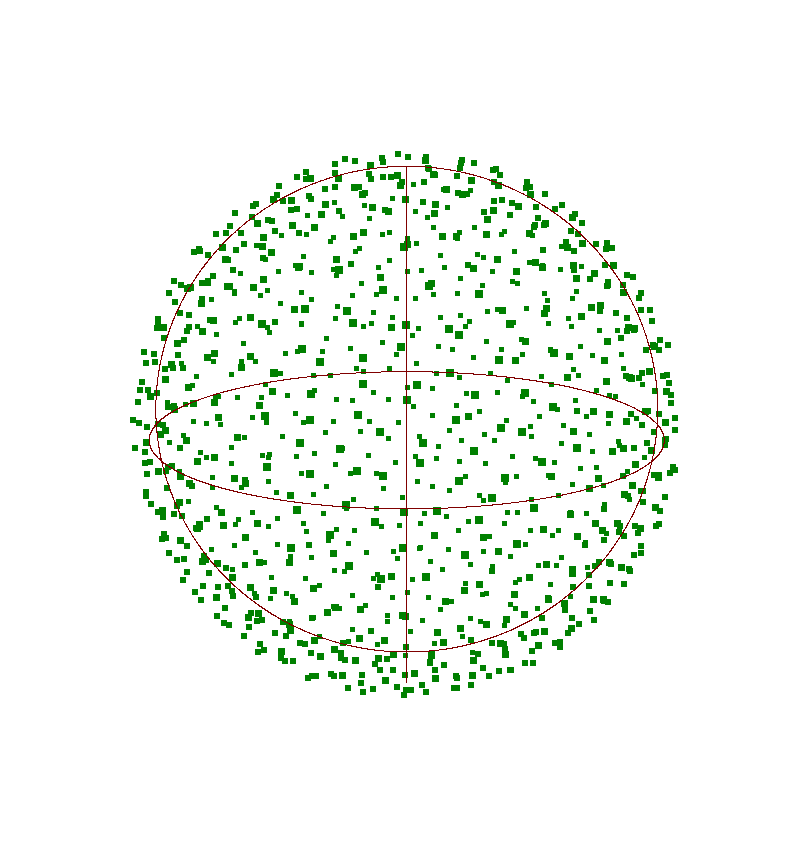}}
  \subfigure[]{\includegraphics[width=.45\linewidth]{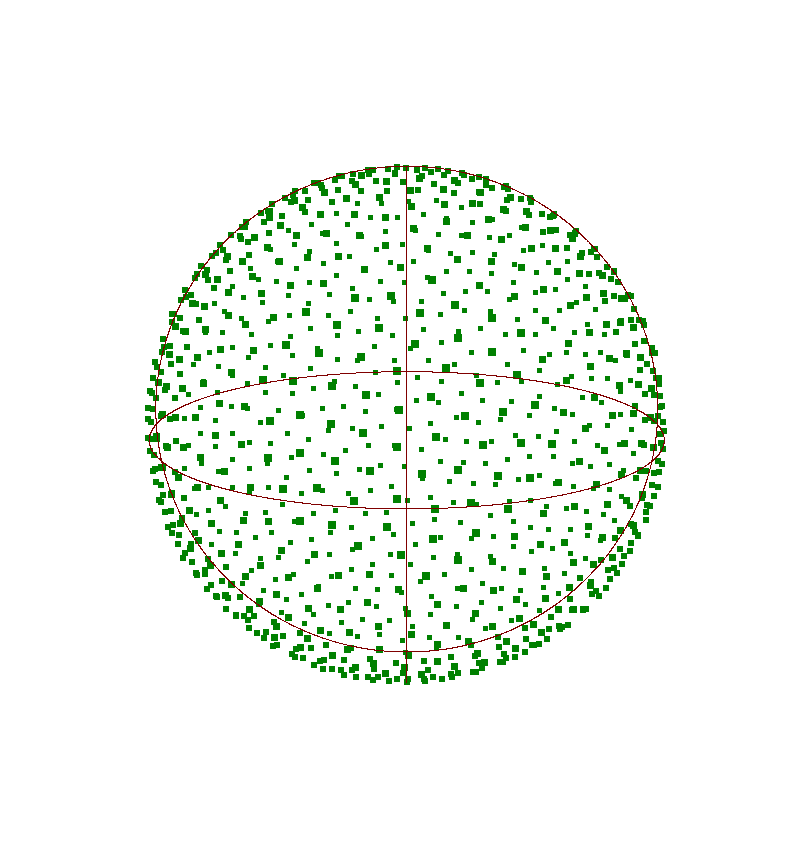}}
  \caption{\label{fig:noisy_sphere}
A uniform sample (a) from a 3D unit sphere (i.e. shown by the `trackball') convolved with Gaussian noise. Presented with this input, an NG with $k=1024$ remains somewhat dispersed for $\lambda = 0$ (b) and $\lambda=6$ (c) while, for $\lambda \ge 10$ (d), the NG disposes itself on the surface of a sphere.}
\end{figure}

\subsection{Interplay between shape and magnification}

\begin{figure}[h]
  \centering
  \subfigure[]{\includegraphics[trim=3 4 3 3,clip,width=.48\linewidth]{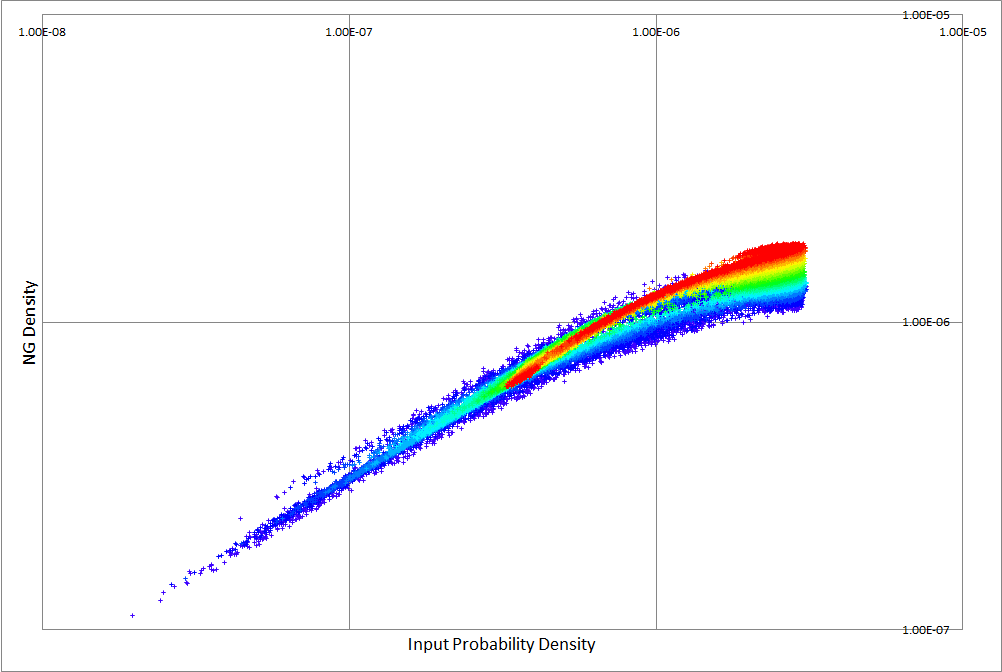}}
  \hspace{.01\linewidth}
  \subfigure[]{\includegraphics[trim=3 4 3 3,clip,width=.48\linewidth]{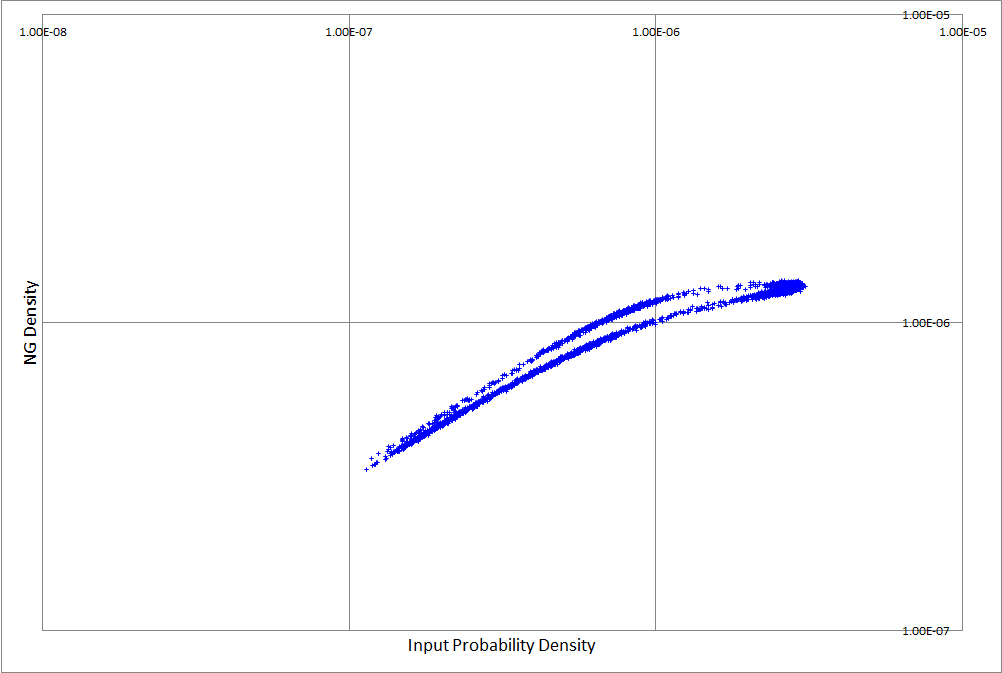}}
  \subfigure[]{\includegraphics[trim=3 4 3 3,clip,width=.48\linewidth]{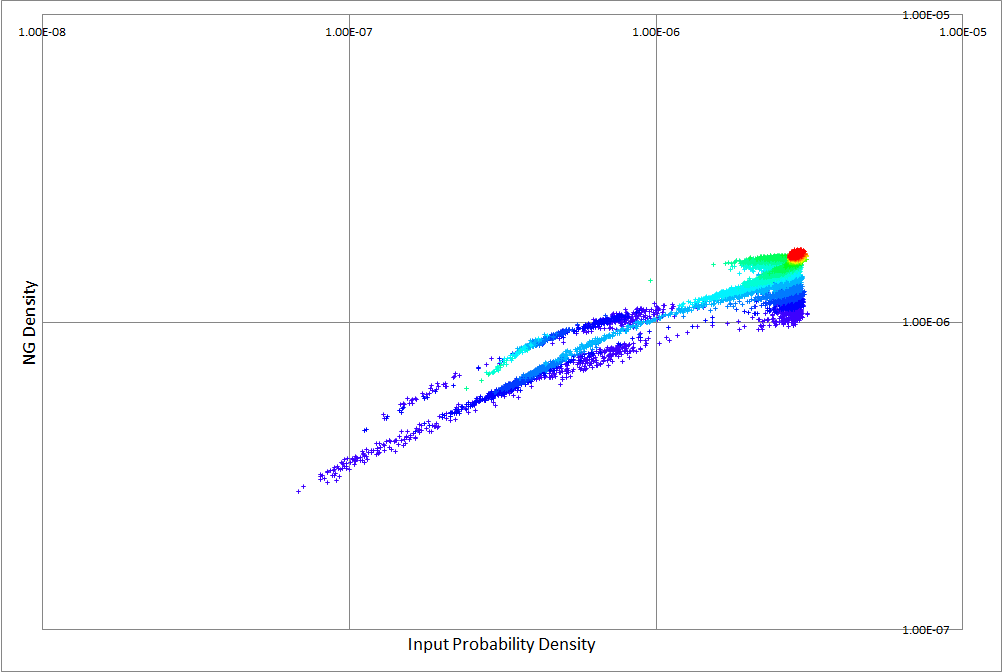}}
  \hspace{.01\linewidth}
  \subfigure[]{\includegraphics[trim=3 4 3 3,clip,width=.48\linewidth]{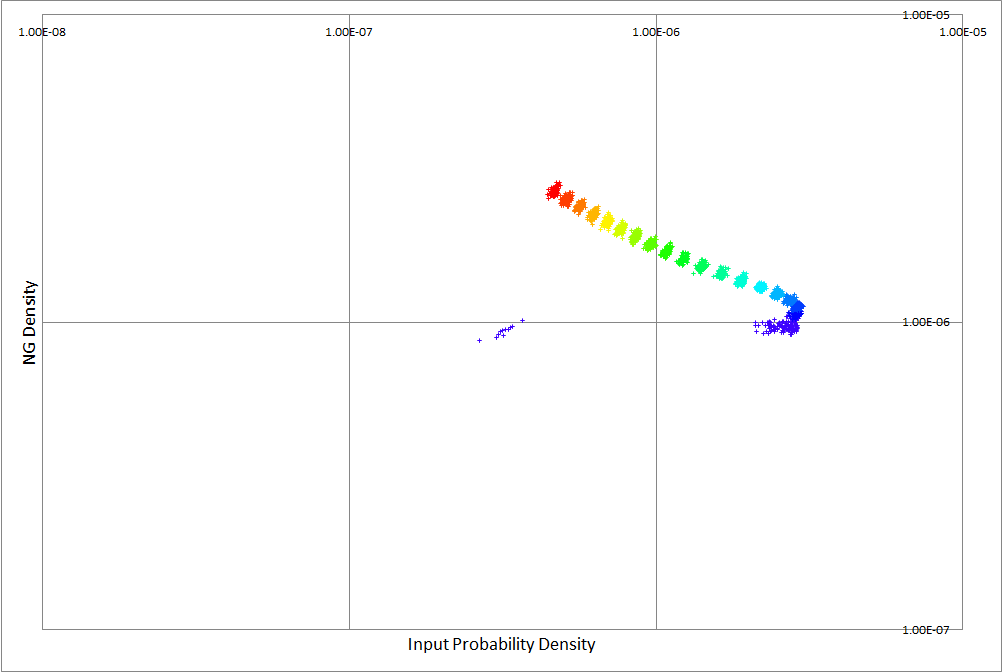}}
  \caption{\label{fig:noisy_sphere_diagrams}
Piecewise justification of the qualitative diagram (see text) describing the NG behavior with the `noisy sphere' in Fig.~\ref{fig:noisy_sphere}. With $k=4096$, the log-log density plot (a) for $\lambda$ values in the range $0$--$18$ shows that the power law is dominant for all units sufficiently away from the sphere while the effects of shape are more pronounced in its proximity; in particular, the detail for $\lambda = 6$ (b) shows the `eye of the needle'. With $k=1024$ and $\lambda \ge 10$ (c) the NG disposes on a sphere, with uniform density (i.e. the red `dot'). With $k=128$ and $\lambda \ge 2$ (d) the NG disposes itself with uniform density on a sphere (blue `spot' on the left) that shrinks progressively, as $\lambda$ increases (in warmer colors).}
\end{figure}

The interplay between the effects produced by shape in data distributions and the power law \eqref{eq:ng_magnification} is effectively exemplified by the behavior shown in Fig.~\ref{fig:noisy_sphere}. The 3D data distribution used in this case is sampled from the convolution between a uniform probability over the unit sphere $\mathbb{S}^2$ and a multivariate normal probability with $\sigma=0.25$ (see \eqref{eq:covariance}). Here, the overall NG behavior can be better discussed with the help of the following qualitative diagram that represents an idealization of the log-log density plots shown in Fig.~\ref{fig:noisy_sphere_diagrams}:
\begin{figure}[H]
  \centering
  \includegraphics[width=.48\linewidth]{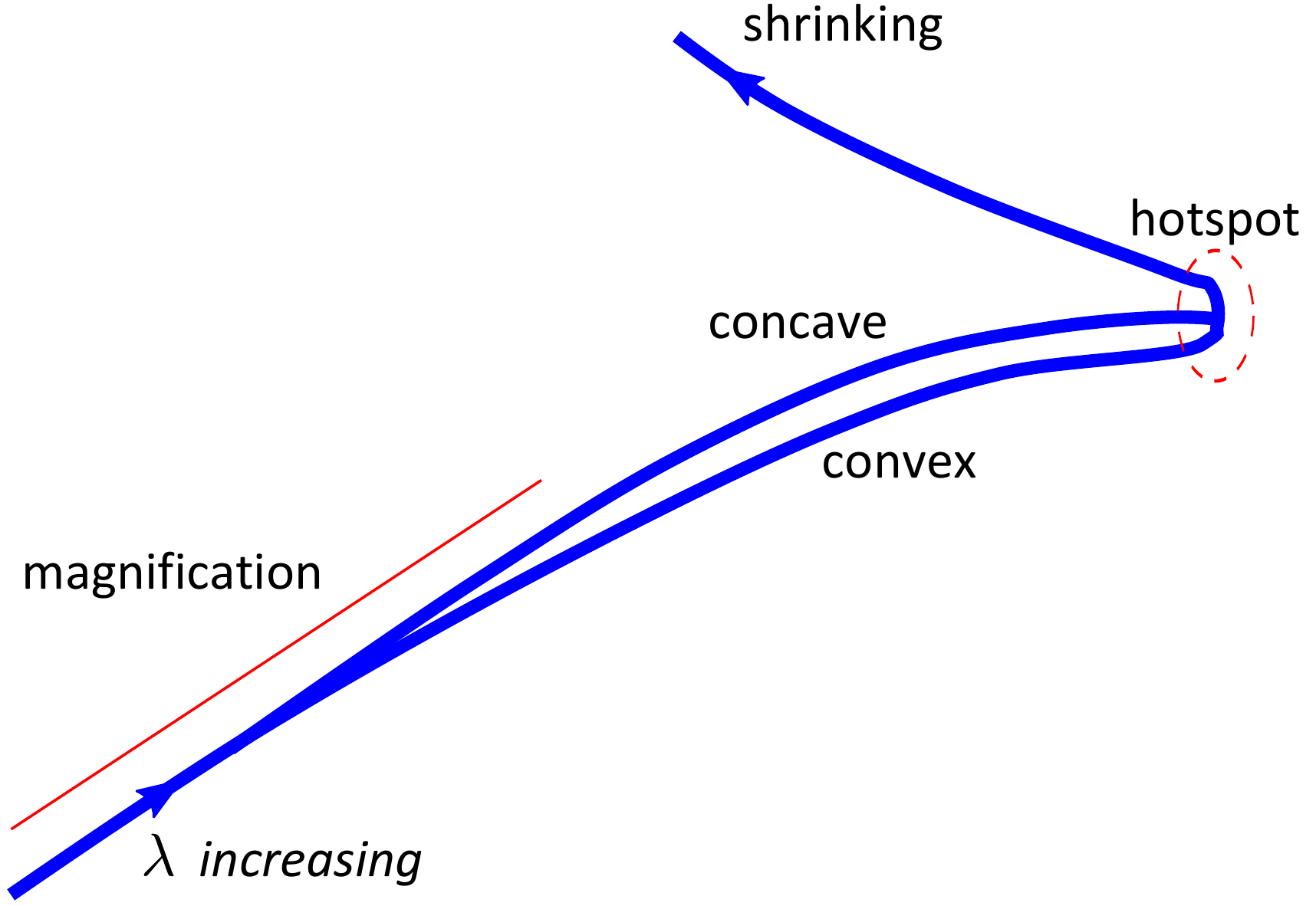}
\end{figure}
When $\lambda$ is low, NG units tend to disperse like in Fig.~\ref{fig:noisy_sphere}(b) and, away from the surface of the sphere, their densities tend to obey the power law whereas, close to the surface, these densities tend to reach a limit value. Overall, on the log-log plot, NG densities occupy a region that goes from `magnification' to `hotspot' in the qualitative diagram, having an extent that also depends on $k$; this region tends to shrink towards the `hotspot' as $\lambda$ increases, as shown in Fig.~\ref{fig:noisy_sphere_diagrams}(a). Interestingly, due to the curvature of the surface, NG densities close to the internal (i.e. `concave') side of the sphere are slightly higher than the densities on the external (i.e. `convex') side; this effect generates the `needle's eye' that can be observed in Fig.~\ref{fig:noisy_sphere_diagrams}(b). When $\lambda$ is sufficiently large, the region occupied by NG densities shrinks to a point-like area (i.e. in the `hostspot'), as shown in Fig.~\ref{fig:noisy_sphere_diagrams}(c), and all NG units dispose in close proximity to the surface of the sphere, as in Fig.~\ref{fig:noisy_sphere}(d). Further increases in the value of $\lambda$ cause the NG to assume the configuration of a sphere with shrinking radius -- while maintaining constant density, as shown in Fig.~\ref{fig:noisy_sphere_diagrams}(d).

\begin{figure}[h]
  \centering
  \subfigure[]{\includegraphics[trim=3 4 3 3,clip,width=.48\linewidth]{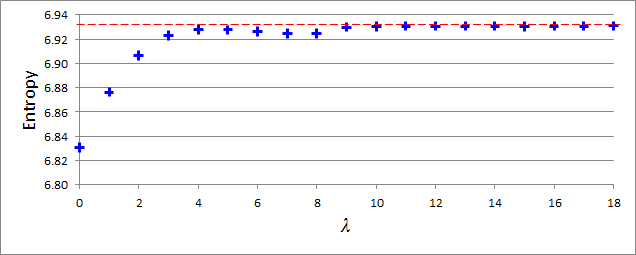}}
  \hspace{.01\linewidth}
  \subfigure[]{\includegraphics[trim=3 4 3 3,clip,width=.48\linewidth]{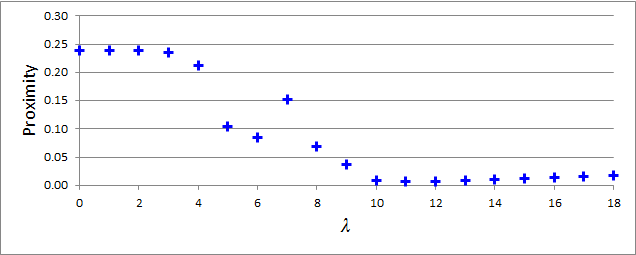}}
  \caption{\label{fig:noisy_sphere_entropy}
Variations in Shannon entropy (a) and proximity function (b) w.r.t. $\lambda$ for final NG configurations in Fig.~\ref{fig:noisy_sphere}. For $\lambda=10$ the value of entropy attains the theoretical maximum $H=6.9314$ (dashed line in red) and remains at the same level even with larger $\lambda$ values. The proximity function (b) $h(\mathbf{W},\mathbb{S}^2)$  reaches a minimum for $\lambda$ in the range 10--13 and then gradually increases as the NG configurations begin to shrink.}
\end{figure}

Fig.~\ref{fig:noisy_sphere_entropy}(a) shows that, with $k=1024$, the Shannon entropy becomes maximal for the same value $\lambda=10$ for which the NG disposes on the sphere, and remains maximal for larger $\lambda$ values. In this case, in fact, the combined effect of spherical shape and constant NG densities makes the Voronoi cells corresponding to NG units to have (almost) the same volume and contain the same probability mass. From another standpoint, Fig.~\ref{fig:noisy_sphere_entropy}(b) describes the NG behavior in terms of \emph{geometrical proximity} to the sphere, as measured by
\begin{equation} 
  \label{eq:proximity}
  h(\mathbf{W},\mathbb{S}^2) := \max_{\mathbf{w}_i \in \mathbf{W}}\left\{\min_{\mathbf{x} \in \mathbb{S}^2}{\| \mathbf{w}_i - \mathbf{x}\| }\right\}.
\end{equation}
As it can be seen, the NG configuration is proximal to the sphere for $\lambda=10$ and then gradually distances itself as $\lambda$ increases, due to the shrinking effect. The NG behavior in point is also remarkably regular: with $k$ values of $128$, $256$, $512$ and $1024$ the minimum of the proximity function occurs when $\lambda$ is approximately equal to $\frac{k}{64}$.

\subsection{More complex shapes}

\begin{figure}[h]
  \centering
  \subfigure[]{\includegraphics[trim=3 4 3 3,clip,width=.38\linewidth]{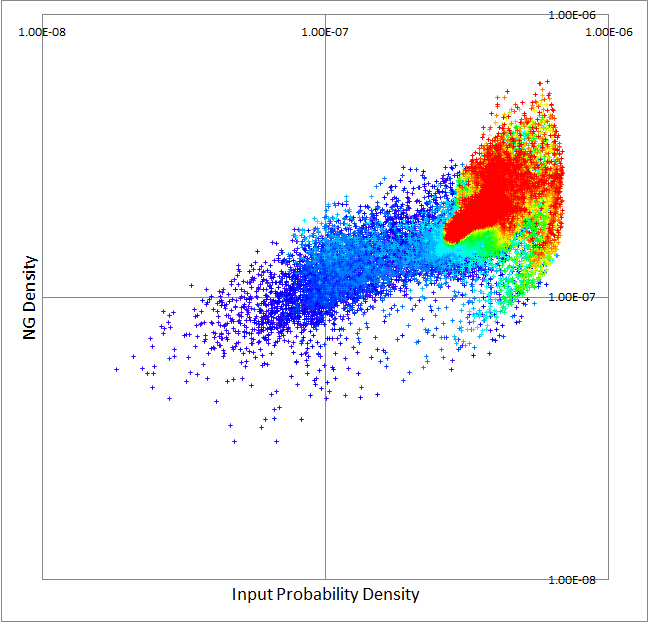}}
  \hspace{.05\linewidth}
  \subfigure[]{\includegraphics[trim=3 4 3 3,clip,width=.38\linewidth]{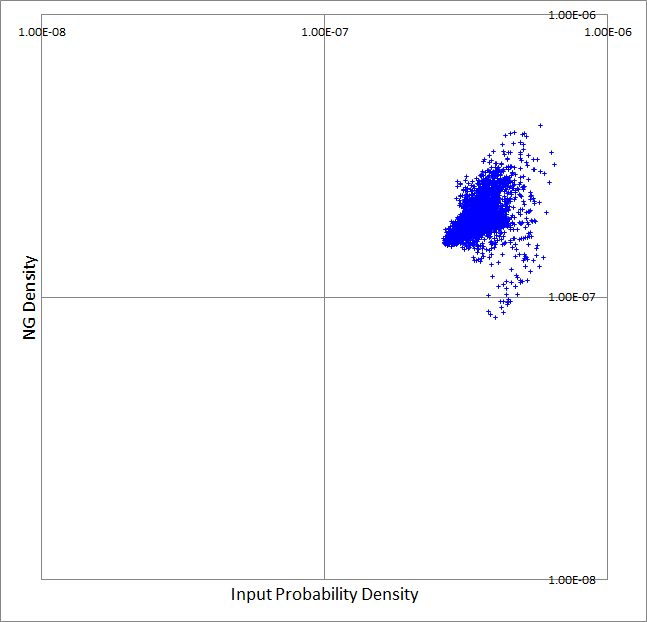}}
  \caption{\label{fig:bunny_diagrams}
With the noisy Stanford bunny and $k=4096$, the log-log density plots (a) for final NG configurations with different values of $\lambda$ in the range 0--18 reflect multiple effects at once: it can be seen however that NG densities tend to be confined in a smaller area as $\lambda$ increases; in particular, the detail (b) shows the densities for the optimal value $\lambda=7$.}
\end{figure}

\begin{figure}[h]
  \centering
  \subfigure[]{\includegraphics[trim=3 4 3 3,clip,width=.48\linewidth]{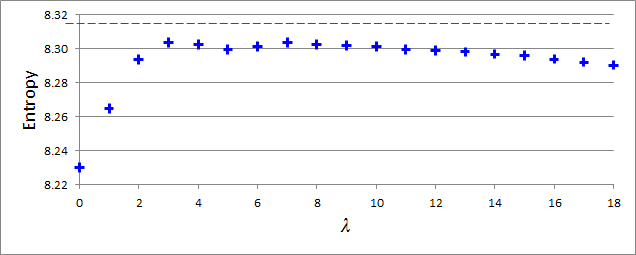}}
  \hspace{.01\linewidth}
  \subfigure[]{\includegraphics[trim=3 4 3 3,clip,width=.48\linewidth]{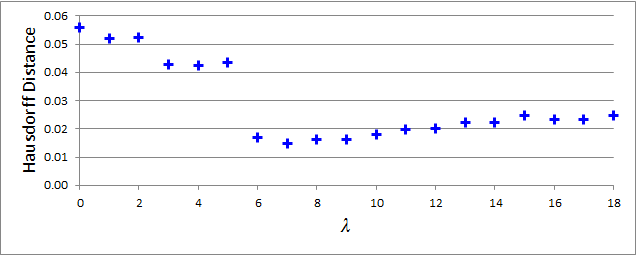}}
  \caption{\label{fig:bunny_entropy}
Variations in Shannon entropy (a) and Hausdorff distance (b) w.r.t. $\lambda$ for the NG experiments with the noisy sample of the Stanford bunny in Fig.~\ref{fig:seq-bunny}(a). The entropy reaches its top value, close to the theoretical maximum (dashed line in red), for $\lambda = 7$ and then decreases with larger $\lambda$ values. The Hausdorff distance (b) reaches a minimum for the same value and then increases gradually.}
\end{figure}

Fig.~\ref{fig:seq-bunny} shows the final NG configuration for an input data distribution obtained by convolving the Stanford bunny point cloud with isotropic gaussian noise with $\sigma = 0.05l$, where $l$ is the side length of the smallest cube containing all the points in the cloud. In the experiment shown, $k=4096$ and $\lambda = 7$. The log-log density plots in Fig.~\ref{fig:bunny_diagrams}(a) show that, as $\lambda$ increases, the NG densities tend to concentrate in a more restricted area. On the other hand, the log-log plots do not show a clear behavior as in the case of the sphere, possibly due to the much richer geometrical features of the bunny, both in terms of curvature and distance between different parts of the surface\footnote{More precisely, the \emph{local feature size} \cite{amenta1999surface} varies greatly over the bunny surface, while it is constant on the sphere.}. Due to the same reason, the Shannon entropy diagram in Fig.~\ref{fig:bunny_entropy}(a) shows a few heights in the range 3--7 and then decreases slowly but steadily; this suggests that with more complex shapes, with different local geometries, there will be no unique and optimal choice for the values of $\lambda$ and $k$. 

Nevertheless, the triangulated shape in Fig.~\ref{fig:seq-bunny}(d) shows the existence, in this case, of a suitable range of $\lambda$ and $k$ values that makes it possible a faithful reconstruction of the original surface from the final NG configuration\footnote{The surface in Fig.~\ref{fig:seq-bunny}(d) has been reconstructed with the \emph{ball pivoting} method implemented in Meshlab \cite{cignoni2008meshlab}.}. A critical parameter for this is the Hausdorff distance, defined as
$$d_H(\mathbf{W},S_ {\text{\emph{bunny}}}) :=  \max\left\{ h(\mathbf{W},S_ {\text{\emph{bunny}}}), h(S_ {\text{\emph{bunny}}}, \mathbf{W})\right\},$$
where $S_ {\text{\emph{bunny}}}$ is the surface of the bunny and $h$ is the proximity function in \eqref{eq:proximity}. A sufficiently low $d_H$, in fact, guarantees the possibility of reconstructing from $\mathbf{W}$ a triangulation that is both \emph{homeomorphic} and geometrically close to the original 2D surface. The diagram in Fig.~\ref{fig:bunny_diagrams}(b) shows that, for the case in point, the minimal $d_H$ value is attained for $\lambda=7$, making this NG configuration the most likely candidate.

\begin{figure}[h]
  \centering
  \subfigure[]{\includegraphics[width=.49\linewidth]{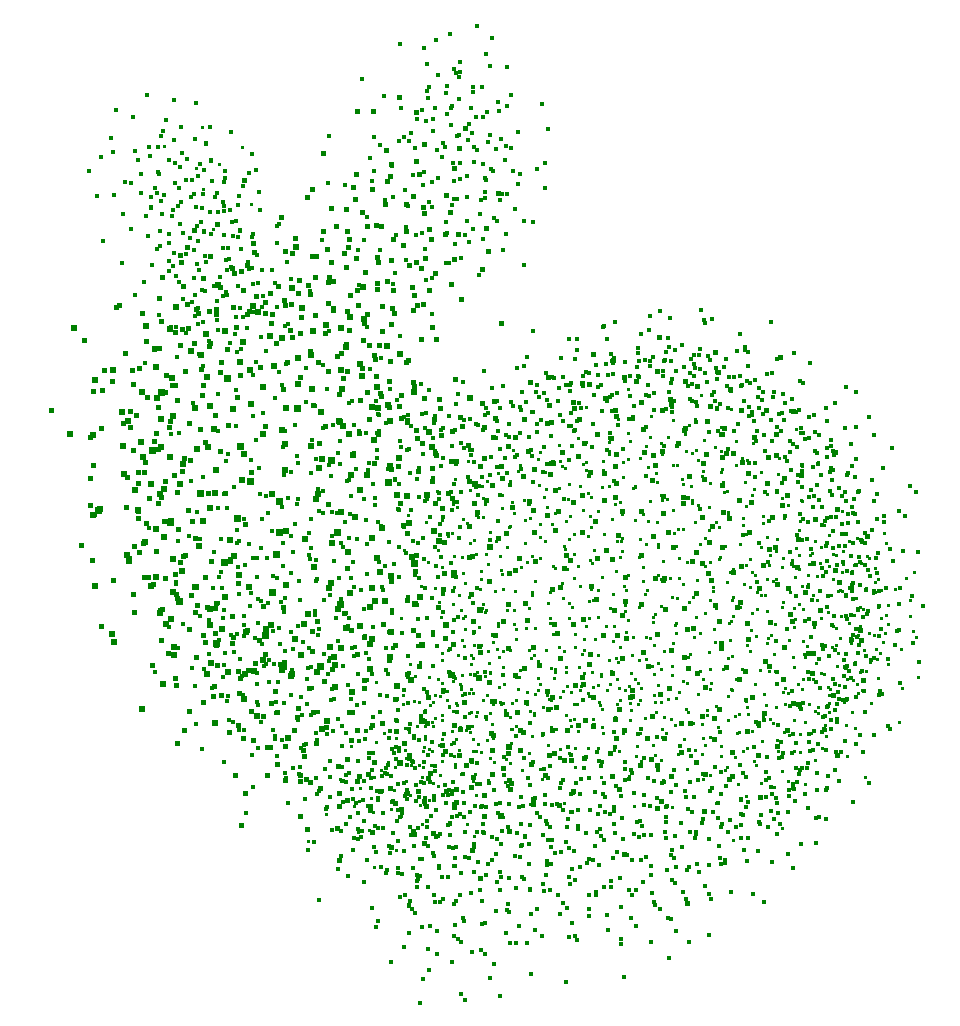}}
  \subfigure[]{\includegraphics[width=.49\linewidth]{bunny-V4096-L7}}
  \subfigure[]{\includegraphics[width=.49\linewidth]{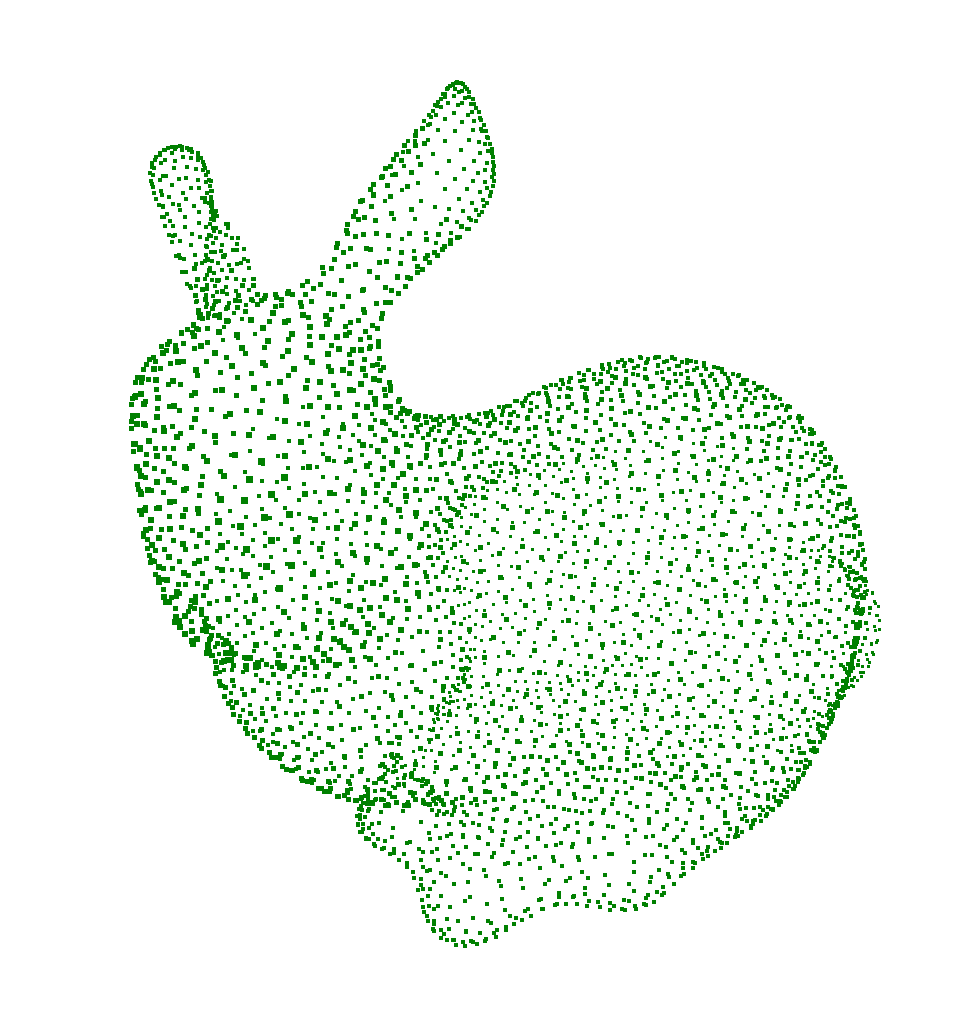}}
  \subfigure[]{\includegraphics[width=.49\linewidth]{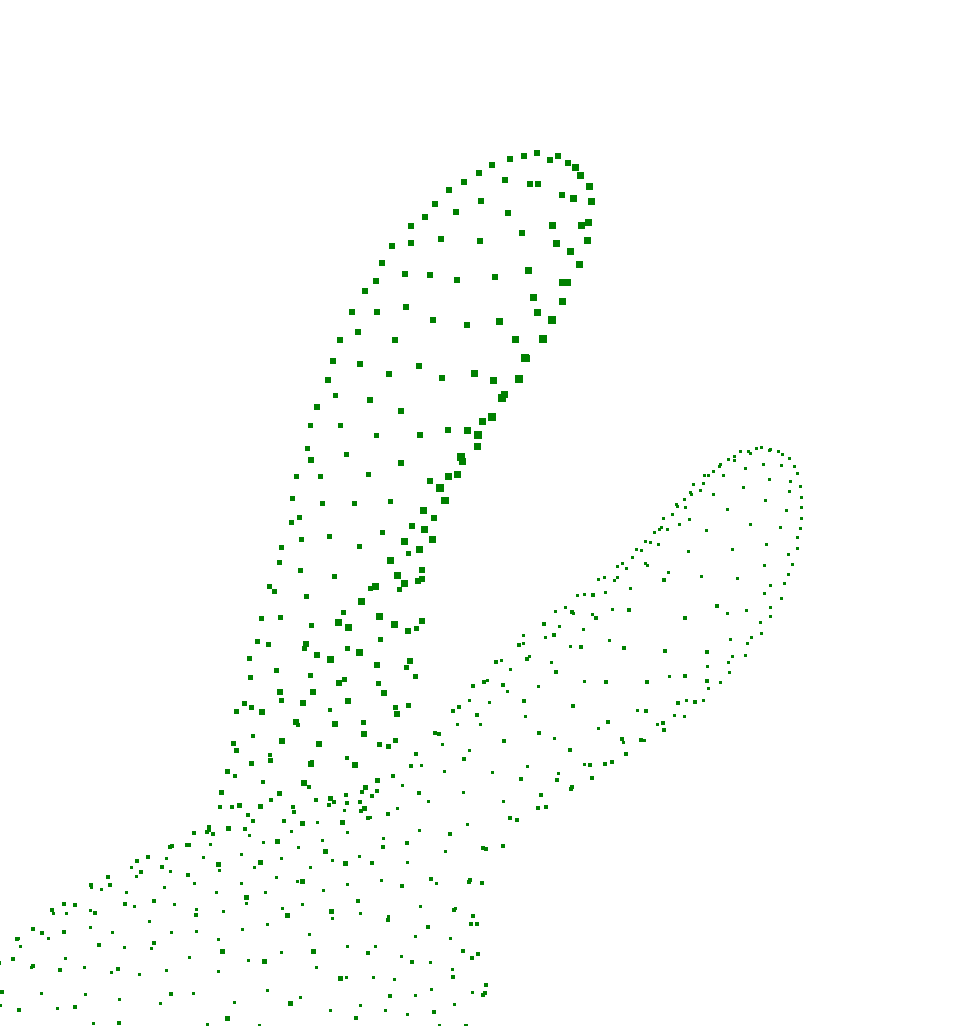}}
  \caption{\label{fig:bunny-lambdas}
Final NG configurations for the noisy sample of the Stanford bunny (see Fig.~\ref{fig:seq-bunny}(b)); with $\lambda=0$ (a) the configuration is too disperse for the purposes of surface reconstruction while with the optimal value $\lambda = 7$ (b) this is possible (see Fig.~\ref{fig:seq-bunny}(d)). With $\lambda = 18$ (c), however, the NG configuration alters significant shape features: as it can be seen in the detail (d), the ears are no longer hollow and the NG disposes on a surface.}
\end{figure}

In general, the suitable range of both $\lambda$ and $k$ values for surface reconstruction will be limited. With the Stanford bunny, for instance, it is not possible to reconstruct the surface for $k\le1024$, as the density of NG units becomes too low in critical areas, like e.g. on the ears of the bunny. Likewise, as shown in Fig.~\ref{fig:bunny-lambdas}, for $\lambda<6$ the final NG configuration is too dispersed and for larger $\lambda$ values the shrinking tendency of the NG entails the loss of important topological details.

On the other hand, this particular NG behavior is quite robust with respect to different (isotropic) noise models. In the experiments performed, the Stanford bunny has been convolved with the sinsuoidal probability in \eqref{eq:sinusoidal} with $r=2\sigma$, where $\sigma = 0.05l$ is the same value as above. With $k=4096$ and $\lambda=7$ the reconstruction is again successful. Further experiments have also been performed by convolving the same point cloud with the uniform probability $\mathcal{U}(B(\mathbf{0},r))$ with $r=\sigma$; for $k=4096$ and $\lambda=6$ the reconstruction is successful as well.

\section{Discussion}
The experimental evidence presented confirms the validity of the \emph{magnification} model \eqref{eq:ng_magnification} for the Neural Gas, even with relatively large values of $\lambda$, whenever the \emph{shape} of the input probability $P$ is not particularly relevant. Vice versa, when the geometrical and topological features of $P$ are more pronounced, the NG behaves in ways that are not explained by the magnification law alone. As already stated, this further evidence is not in contrast with the original NG model in \cite{Martinetz-etal93} as its analytical derivation relies in fact on neutralizing the effects of shape.

The experiments performed show that the effects of shape in $P$, when present, become dominant in determining the final NG configurations. Although at present these phenomena can be assessed only empirically, the NG response to geometric features seems to be describable in precise terms, at least with specific and controlled benchmark shapes.  

For practical purposes, the experiments also show that the NG behavior described has the interesting potential for performing a specific sort of \emph{online blind deconvolution} \cite{hirsch2011online}; \emph{online} as the NG adapts its configuration via SGD in response to a stream of incoming pointwise signals, and \emph{blind} since the noise model needs not be known beforehand. 

Further work is clearly necessary to assess the usefulness of the NG behavior with more realistic and complex examples, for instance with the non-uniform sampling of shapes and/or with anisotropic noise models. In addition, another direction for further studies is about the impact on the NG response to shape with the various techniques for \emph{magnification control} described in the literature \cite{jain2004forbidden, claussen2005magnification, villmann2006magnification, merenyi2007explicit, hammer2007magnification}, i.e. having the objective of controlling the $\alpha$ factor in the power law \eqref{eq:ng_magnification}.

\section*{Acknowledgment}

This work is funded by the national governments and the European Union through the ENIAC JU project DeNeCoR under grant agreement number 32425.
The second author gratefully acknowledges support by the Italian FIRB ``Futuro in Ricerca'' grant RBFR10DGUA.



%

\bibliographystyle{IEEEtran}
\bibliography{ng}




\end{document}